\documentclass[smallextended]{preprint-svjour3}       
\RequirePackage{fix-cm}
\smartqed  
\usepackage{natbib}

\usepackage[utf8]{inputenc} 
\usepackage[T1]{fontenc}    
\usepackage{enumitem}
\usepackage{amsmath}
\usepackage{color}
\usepackage[toc,page]{appendix}
\usepackage{amssymb}
\usepackage{graphicx}
\usepackage{epstopdf}
\usepackage{hyperref}
\usepackage{alltt}
\usepackage{listings}
\usepackage{array}
\usepackage{caption}
\usepackage{subcaption}
\usepackage{sparklines}
\usepackage{geometry}

\newcommand{\secref}[1]{Section~\ref{#1}}

\newcommand{\tblref}[1]{Table~\ref{#1}}
\newcommand{\figref}[1]{Figure~\ref{#1}}

\newcommand{\cut}[1]{}
\newcommand{\todo}[1]{{{\small\color{red}{[#1]}}}}
\renewcommand{\slash}{\texttt{\char`\\}}

\DeclareMathOperator{\Ex}{\mathbb{E}}

\newcommand{\Imp}{\fontfamily{cmr}\textsc{Impact}}
\newcommand{\Impo}{\fontfamily{cmr}\textsc{Import}}
\newcommand{\CImp}{\fontfamily{cmr}\textsc{CatImpact}}
\newcommand{\CImpo}{\fontfamily{cmr}\textsc{CatImport}}
\newcommand{\simp}{\fontfamily{cmr}\textsc{\small StratImpact}}

\newcommand{\spd}{\fontfamily{cmr}\textsc{\small StratPD}}
\newcommand{\cspd}{\fontfamily{cmr}\textsc{\small CatStratPD}}

\renewcommand{\xi}{x^{(i)}}

\setlist[enumerate]{itemsep=-1mm}

\journalname{PREPRINT}

\begin{document}

\title{Nonparametric Feature Impact and Importance}

\cut{
\author{Terence Parr \email parrt@cs.usfca.edu
\addr University of San Francisco\\
\AND James D. Wilson \email jdwilson4@usfca.edu
\addr University of San Francisco\\
\AND Jeff Hamrick \email jhamrick@usfca.edu
      \addr University of San Francisco}
}

\author{Terence Parr \and James D. Wilson* \and Jeff Hamrick}
\institute{Terence Parr \at
  University of San Francisco, 
  \email{{\tt parrt@cs.usfca.edu}}
  \and
  James D. Wilson \at
  University of San Francisco
  \email{{\tt jdwilson4@usfca.edu}}
  \and
  Jeff Hamrick \at
  University of San Francisco
  \email{{\tt jhamrick@usfca.edu}} 
}


\maketitle

\begin{abstract}%
Practitioners use feature importance to rank and eliminate weak predictors during model development in an effort to simplify models and improve generality.  Unfortunately, they also routinely conflate such feature importance measures with feature impact, the isolated effect of an explanatory variable on the response variable.   This can lead to real-world consequences when importance is inappropriately interpreted as impact for business or medical insight purposes. The dominant approach for computing importances is through interrogation of a fitted model, which works well for feature selection, but gives distorted measures of feature impact. The same method applied to the same data set can yield different feature importances, depending on the model, leading us to conclude that impact should be computed directly from the data.  While there are nonparametric feature selection algorithms, they typically provide feature rankings, rather than measures of impact or importance. They also typically focus on single-variable associations with the response. In this paper, we give mathematical definitions of feature impact and importance, derived from partial dependence curves, that operate directly on the data. To assess quality, we show that features ranked by these definitions are competitive with existing feature selection techniques using three real data sets for predictive tasks.
\end{abstract}


\section{Introduction}
\label{sec:intro}

Among data analysis techniques, feature importance is one of the most widely applied and practitioners use it for two key purposes: (1) to select features for predictive models, dropping the least predictive features to simplify and potentially increase the generality of the model and (2) to gain business, medical, or other insights, such as product characteristics valued by customers or treatments contributing to patient recovery.  To distinguish the two use cases, we will refer to feature predictiveness for modeling purposes as {\em importance} (the usual meaning) and the effect of features on business or medical response variables as {\em impact}.

While some feature importance approaches work directly on the data, such as minimal-redundancy-maximal-relevance (mRMR) by \cite{mRMR}, almost all algorithms used in practice rank features by interrogating a fitted model provided by the user.  Examples include permutation importance by \cite{RF}, drop column importance, and SHAP by \cite{shap}; LIME by \cite{lime} interrogates subsidiary models to analyze such fitted models. It is accepted as self-evident that identifying the most predictive features for a model is best done through interrogation of that  model, but this is not always the case.  For example, when asked to identify the single most important feature of a real dataset \citep{bulldozer} for a random forest (RF), the features selected by model-based techniques get twice the validation error of the nonparametric technique proposed in this paper; see \figref{fig:topk}c. Still, model interrogation is generally very effective in practice for feature importance purposes.

Feature importance should not, however, be interpreted as feature impact for several reasons. First, predictive features do not always coincide with impactful features; e.g., models unable to capture complex nonlinear feature-response relationships rank such features as unimportant, even if they have large impacts on the response. Next, practitioners must develop models accurate enough to yield meaningful feature importances, but there is no definition of ``accurate enough.'' Finally, it is possible to get very different feature importances (and hence impacts) running the same algorithm on the same data, just by choosing a different model. This is despite the fact that feature impacts are relationships that exist in the data, with or without a model.

\begin{figure}[htbp]
\begin{center}
\includegraphics[scale=0.58]{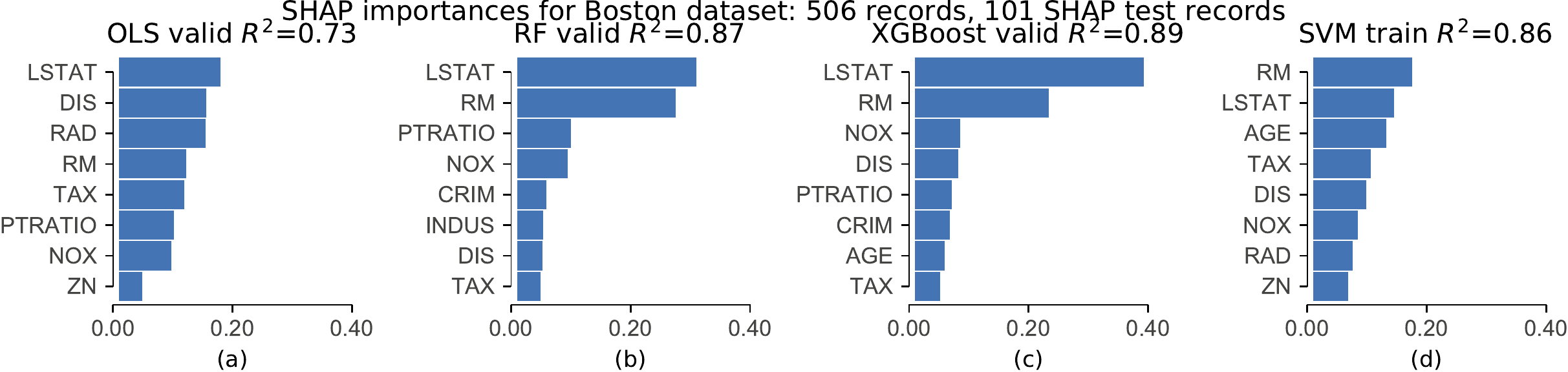}
\vspace{-3mm}
\caption{\small Ranking and relative predictiveness of the top 8 of 13 Boston data set features determined by SHAP interrogating four different models.  There is considerable variation between plots even in the two most important features. For example, while RF and XGBoost rank {\tt LSTAT} and {\tt RM} first but the RF gives {\tt RM} much more weight. The SVM reverses that ranking. Model hyperparameters were tuned with 5-fold cross-validation grid search on a variety of hyperparameters over an 80\% training set.  SHAP explains the 20\% validation set.}
\label{fig:diff-models}
\end{center}
\end{figure}

Consider the feature importance charts in \figref{fig:diff-models} derived from four different models on the same well-known Boston toy data set, as computed by SHAP. The linear model (a) struggles to capture the relationship between features and response variable (validation $R^2$=0.73), so those importances are less trustworthy.  In contrast, the (b) RF, (c) boosted trees, and (d) support vector machine (SVM) models capture the relationship well (each with $R^2 > 0.85$). The problem is that SHAP derives meaningfully different feature importances from each model, as most model-based techniques would. The differences arise because feature impact is distorted by the lens' of the models (yielding importances). The differences might be appropriate for model feature selection, but it is unclear which ranking, if any, gives the feature impacts. 

Unfortunately, practitioners routinely conflate model-based feature importance with impact and have, consequently, likely made business or medical decisions based upon faulty information. Despite the potentially serious real-world consequences resulting from inappropriate application of importances, research attention has focused primarily on feature importance rather than feature impact. 

In this paper, we address this deficiency by contributing (1) a straightforward mathematical formula as an ideal for computing feature impact, and related feature importance, that does not require predictions from a fitted model and (2) a prototype implementation called \simp{} that yields plausible feature impacts. By not requiring predictions from a model, we also make feature impact methods accessible to the vast  community of business analysts and scientists that lack the expertise to choose, tune, and evaluate models. To assess \simp{} quality, we use importance as a proxy to show that it is competitive on real data with existing importance techniques, as measured by validation errors on models trained using the top $k$ feature importances.  

We measure feature impact as a function of its partial dependence curve, as \cite{pdvim} did, because partial dependences (ideally) isolate the effect of a single variable on the response. In contrast to \cite{pdvim}, we use a nonparametric method called \spd{} (\citealt{stratpd}) to estimate partial dependences without predictions from a fitted model, which allows us to compute feature impact not just feature importance. \spd{} also isolates partial dependence curves in the presence of strong codependencies between features. The partial dependence approach is flexible in that it opens up the possibility of computing importances using any partial dependence method, such as Friedman's original definition (\citealt{PDP}) and ALE (\citealt{ALE}). (We will refer to Friedman's original definition as FPD to distinguish it from the general notion of partial dependence.) SHAP also fits into this perspective since the average SHAP value at any single feature value forms a point on a mean-centered partial dependence curve. Our prototype is currently limited to regression but accepts numerical and label-encoded categorical explanatory variables; a similar approach should work for classification. The software is available via Python package {\tt stratx} with source at {\tt github.com/parrt/stratx}. 

We begin by giving definitions of feature impact and importance in \secref{sec:def}, then survey existing nonparametric and model-dependent techniques in \secref{sec:existing}. \secref{sec:experiments} assesses the quality of \simp{} importance values by examining how well they rank model features in terms of predictiveness. We finish in \secref{sec:discussion} with a discussion of the proposed technique's effectiveness and future work.

\section{Definitions of impact and importance}\label{sec:def}

\cut{Practitioners loosely define feature importance as feature predictiveness, which presupposes a fitted predictive model, probably because importances are so often used for feature selection during model development.  Research  focuses on more accurately identifying the impact of features upon model predictions.  But, relying on a fitted model makes it difficult to tease apart the true feature importance from the ability of the model to exploit that feature for prediction purposes. Rather than measuring feature impact on {\em model predictions}, we propose avoiding the model completely to define feature importance as the average impact of a feature on the {\em data set response values}.}

In special circumstances, we know the precise impact of each feature $x_j$. Assume we are given the training data pair ($\bf X, y$) where ${\bf X} = [x^{(1)}, \ldots, x^{(n)}]$ is an $n \times p$ matrix whose $p$ columns represent observed features and ${\bf y}$ is the $n \times 1$ vector of responses; ${\bf X}_j$ is the $n \times 1$ column of data associated with feature $x_j$.  If a data set is generated using a linear function, $y = \beta_0 + \sum_{j=1}^p \beta_j x_j$, then coefficient $\beta_j$ corresponds  to the impact of $x_j$ for $j=1,..,p$.  $\beta_j$ is the impact on $y$ for a unit change in feature $x_j$, holding other features constant.

To hold features constant for any smooth and continuous generator function $f:\mathbb{R}^{p} \rightarrow \mathbb{R}$ that precisely maps each $\xi$ to $y^{(i)}$, ${y^{(i)}} = f(\xi)$, we can take the partial derivatives of $f$ with respect to each feature $x_j$; e.g., for linear functions, ${\partial y}/{\partial x_j}=\beta_j$. Integrating the partial derivative then gives the {\em idealized partial dependence} (\citealt{stratpd}) of $y$ on $x_j$, the isolated contribution of $x_j$ at $z$ to $y$:

\begin{equation}\label{eq:pd}
\text{\it PD}_j(x_j = z) = \int_{min(x_j)}^z \frac{\partial f}{\partial x_j} dx_j
\end{equation}

Using partial derivatives to isolate the effect of variables on the response was used prior to \spd{} by ALE (\citealt{ALE}) and {\em Integrated Gradients} (IG) (\citealt{intgrad}). The key distinction is that  \spd{} integrates over the derivative of the generator function, ${\partial f}/{\partial x_j}$, estimated from the raw training data, whereas, previous techniques integrate over the derivative of model $\hat{f}$ that estimates $f$: ${\partial \hat{f}}/{\partial x_j}$. While there are advantages to using models, such as their ability to smooth over noise, properly choosing and tuning a machine learning model presents a barrier to many user communities, such as business analysts, scientists, and medical researchers. Further, without potential distortions from a model, \spd{} supports the measurement of impact, not just importance.

\subsection{Impact and importance for numerical features}

To go from the idealized partial dependence of $x_j$ to feature impact, we assume that the larger the ``mass'' under $x_j$'s partial dependence curve, the larger $x_j$'s impact on $y$.

~\\
\noindent {\bf Definition 1} The (non-normalized) {\em nonparametric feature impact} of $x_j$ is the area under the magnitude of $x_j$'s idealized partial dependence:

\begin{equation}
\Imp_j = \int_{\min({\bf X}_j)}^{\max({\bf X}_j)} |PD_j(x_j)| dx_j
\end{equation}

\noindent In practice, we approximate the integral with a Riemann sum of rectangular regions:

\begin{equation}
\Imp_j \approx \sum_{x_j \in \{{\bf X}_j\}} |PD_j(x_j)| \Delta x_j
\end{equation}

\noindent  where $\{{\bf X}_j\}$ is the set of unique ${\bf X}_j$ values ($\{{\bf X}_j^{(i)}\}_{i=1..n}$) and $n_j$ is the number of unique ${\bf X}_j$. 
The usual definition of region width, $\Delta x_j = {(\max({\bf X}_j) - \min({\bf X}_j))}/{n_j}$, is inappropriate in practice because ${\bf X}_j$ often has large gaps in $x_j$ space and impact units would include $x_j$'s units (e.g., $rent \times bedrooms$ or $rent \times hasparking$). That would make impact scores incomparable across features. By defining $\Delta x_j = 1/n_j$, the fraction of unique ${\bf X}_j$ values covered by one $PD_j$ value, widths are not skewed by empty $x_j$ gaps and impact units become those of $y$. That reduces $x_j$'s impact estimate to the average magnitude of $PD_j$:

\noindent This formula works for any partial dependence curve, either by examining output from a fitted model, $\hat{f}$, or by estimating the partial derivative of $f$ directly from the data and then integrating, as \spd{} does. Dividing $x_j$'s impact by the sum of all impacts, converts impact units from $y$'s units to [0,1]:

\[
\Imp_j \approx  \sum_{x_j \in \{{\bf X}_j\}} |PD_j(x_j)| \times \frac{1}{n_j} = \overline{|PD_j|}
\]

~\\
\noindent {\bf Definition 2} The {\em normalized nonparametric feature impact} of $x_j$ is the ratio of the average magnitude of $x_j$'s partial dependence to the total for all variables:

\begin{equation}\label{eq:Epd2a}
\Imp_j^{*} = \frac{\overline{|\text{\it PD}_j|}}{\sum_{k=1}^p \overline{|\text{\it PD}_k|}}
\end{equation}

Intuitively, the impact of $x_j$ is how much, on average, the values of $x_j$ are expected to push $y$ away from a zero baseline. We deliberately chose this  definition instead of measuring how much $x_j$ pushes $y$ away from the average response, $\overline{y}$, as SHAP does.  The average response includes the effects of all $x_j$, which hinders isolation of individual feature impacts. For example, the impact of $x_1$ on quadratic $y = x_1^2+x_2+100$ is $x_1^2$, not $|\overline{y} - x_1^2|$.  

\begin{figure}
\centering
\begin{minipage}{.56\textwidth}
  \centering
\includegraphics[scale=0.57]{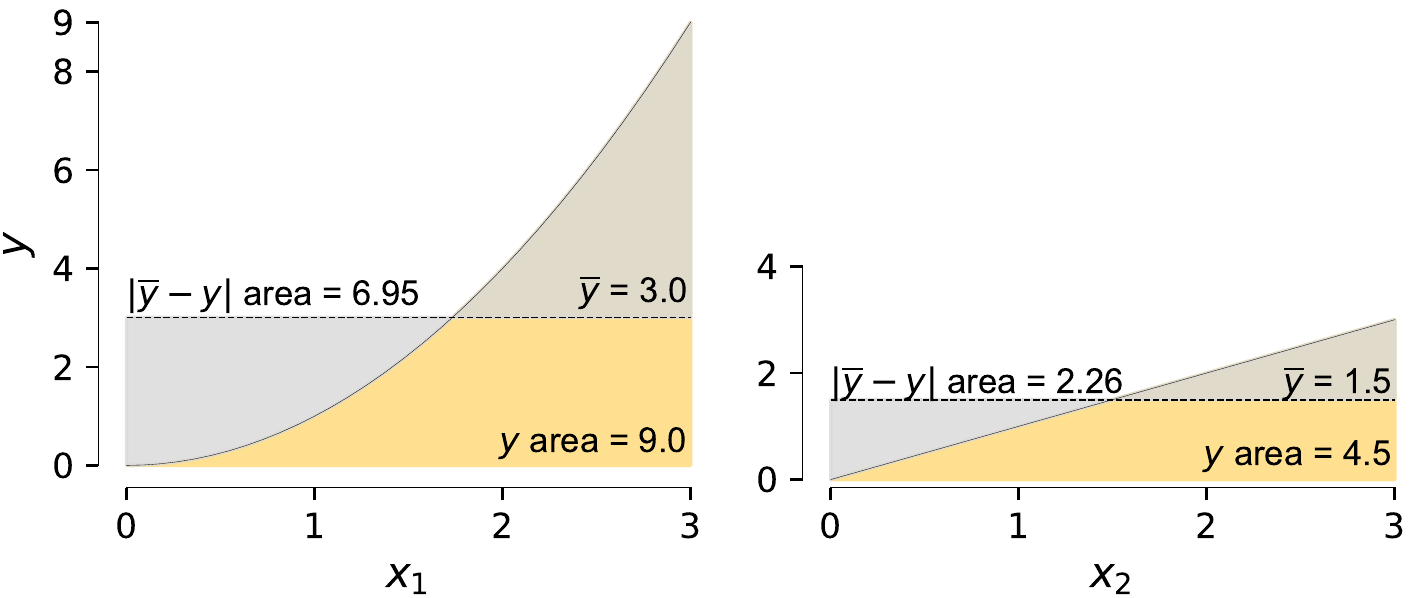}
\vspace{-3mm}
  \captionof{figure}{\small The area under $x_1$ and $x_2$ PD curves represent $\simp_1$, $\simp_2$ for $y = x_1^2 + x_2 + 100$ in range $[0,3]$. Compare the areas straddling the means and the areas under the partial dependence curves.}
\label{fig:quad-area}
\end{minipage}
\hfill
\begin{minipage}{.41\textwidth}
  \centering
\includegraphics[scale=0.5]{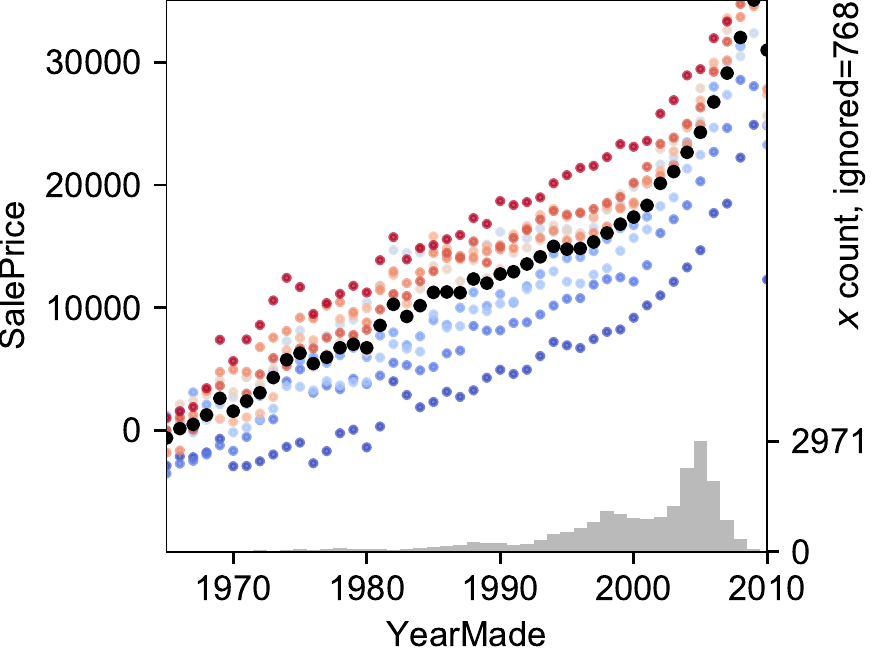}  \vspace{-3mm}
\captionof{figure}{\small \simp{} curve for bulldozer {\tt SalePrice} on {\tt YearMade} including the histogram used to weight the partial dependence to obtain feature importances. A 20k sample of all 363k records; \textasciitilde0.5\% of samples stratified into regions with single {\tt YearMade} values.}
\label{fig:yearmade}
\end{minipage}
\end{figure}

\figref{fig:quad-area} illustrates how the area under the $x_1^2$ and $x_2$ PD curves differ from the area straddling the means for $x_1, x_2 \sim U(0,3)$. The area under-the-curve ratio of $x_1$-to-$x_2$ is 2-to-1 (9/4.5), whereas the ratio of the area straddling the mean has roughly a 3-to-1 ratio (6.95/2.26).  The symbolic partial derivatives are ${\partial y}/{\partial x_1} = 2 x_1$ and ${\partial y}/{\partial x_2} = 1$, so $\text{\it PD}_1 = x_1^2$ and $\text{\it PD}_2 = x_2$. Integrating those gives $\Imp_1 = \int_0^3 x_1^2 dx_1 = \frac{x_1^3}{3} \big |_0^3 = 9$ and $\Imp_2 = \int_0^3 x_2^2 dx_2 = \frac{x_2^2}{2} \big |_0^3 = 4.5$; $\Imp_1^{*} = 0.\overline{66}$ and $\Imp_2^{*} = 0.\overline{33}$.

The $PD_j$ curve represents how $x_j$ effects $y$, but does not take into consideration the distribution of $x_j$ in ${\bf X}_j$.  Consider the \spd{} curve and $x_j$ histogram in \figref{fig:yearmade} for feature {\tt\small YearMade} from the bulldozer auction data set \citep{bulldozer}. The colored curves represent ten bootstraps from the same training set and the black dots give the average curve.   From a business perspective, knowing that increases in bulldozer age continue to reduce price is useful, but new bulldozers will represent the bulk of any model validation set.  Because model feature selection is often assessed using  validation error, feature selection is sensitive to the distribution of $x_j$. This suggests that feature importance should take into consideration the distribution of $x_j$, leading to the following importance definition:

~\\
\noindent {\bf Definition 3} The {\em normalized nonparametric feature importance} of $x_j$ is the ratio of $x_j$'s average partial dependence magnitude, weighted by $x_j$'s distribution, to the total of all weighted averages:


\begin{equation}\label{eq:Epd3b}
\Impo_j^{*} = \frac{\Impo_j}{\sum_{k=1}^p \Impo_k}
\end{equation}

\noindent where

\begin{equation}\label{eq:Epd3c}
\Impo_j \approx \sum_{x_j \in \{{\bf X}_j\}} \frac{n_{x_j}}{n} \times |PD_j(x_j)|
\end{equation}

\noindent and $n_{x_j}$ is the number of $x_j$ values in ${\bf X}_j$; i.e., $(x_j, n_{x_j})_{x_j \in \{{\bf X}_j\}}$ is the histogram of ${\bf X}_j$.


Measuring impact as the average partial dependence generalizes to non-ordinal categorical explanatory features encoded as unique integers, with a small modification.  

\subsection{Impact and importance for categorical features}

Partial dependence curves for numerical features use the leftmost $x_j$ value as the zero reference point, but nominal categorical features have no meaningful order. That implies we can choose any category as the zero reference category, which shifts the partial dependence plot up or down, but does not alter the relative $y$ values among the category levels. For example, consider relative $y$ values for four categories (0, 1, 1, 1) where the first category is the reference.  Choosing the second category as the reference yields relative $y$ values (-1, 0, 0, 0). The impacts (average magnitude) for these two variations are 3/4 versus 1/4, respectively, but the choice of reference category should not affect the impact metric computed on the same partial dependence data. Worse, picking a category level whose $y$ is an outlier strongly biases the impact because the outlier pushes up (or down) all category $y$ values by a biased amount.  Instead of picking a specific level as the reference, therefore, we use that feature's partial dependence average value as the reference zero:\\

\noindent {\bf Definition 4} The {\em normalized nonparametric categorical feature impact} of $x_j$ is the ratio of the average magnitude of $x_j$'s mean-centered partial dependence to the total for all variables:

\begin{equation}\label{eq:Cpd4a}
\CImp_j^{*} = \frac{\overline{|\text{\it PD}_j - \overline{\text{\it PD}_j}|}}{\sum_{k=1}^p \overline{|\text{\it PD}_k - \overline{\text{\it PD}_j}|}}
\end{equation}

~\\

\noindent The definition of categorical variable importance mirrors the definition for numerical variables, which weights impact by the distribution of the $x_j$'s:

~\\

\noindent {\bf Definition 5} The {\em normalized nonparametric categorical feature importance} of $x_j$ is the ratio of $x_j$'s expected mean-centered partial dependence magnitude to the total of all:

\begin{equation}\label{eq:Epd2b}
\CImpo_j^{*} = \frac{\CImpo_j}{\sum_{k=1}^p \CImpo_k}
\end{equation}

\noindent where

\begin{equation}\label{eq:Epd2c}
\CImpo_j \approx \sum_{x_j \in \{{\bf X}_j\}} \frac{n_{x_j}}{n} \times |PD_j(x_j)- \overline{\text{\it PD}_j}|
\end{equation}

By choosing $\overline{\text{\it PD}_j}$ as the reference value (instead of a specific category's $\text{\it PD}_j$ value), our definition moves closer to SHAP's mean-centered approach, which we disagreed with above, but only out of necessity for categorical variables. A key difference is that $\CImpo_j$ centers on $\overline{\text{\it PD}_j}$ rather than the overall average, $\bar{y}$, that includes the effect of all features. With these definitions in mind, we make a more detailed comparison to related work in the next section.

\section{Existing methods}\label{sec:existing}

In this paper, we are primarily concerned with identifying the most impactful features, such as needed in business or medical applications. But, because virtually all related research focuses on feature importance and, because practitioners commonly assume feature importance is the same as feature impact, it is appropriate to compare \simp{} to  feature importance methods. Feature importance methods for labeled data sets (with both $\bf X$ and $\bf y$) are broadly categorized into data analysis and model analysis techniques, sometimes called {\em filter} and {\em wrapper} methods \citep{tsanas}. Data analysis techniques analyze the data directly to identify important features, whereas model analysis techniques rely on predictions from fitted models.

\subsection{Data analysis techniques}

The simplest technique to identify important or relevant regression features is to rank them by their Spearman's rank correlation coefficient \citep{spearmans}; the feature with the largest coefficient is taken to be the most important. This method works well for independent features, but suffers in the presence of codependent features.   Groups of features with similar relationships to the response variable receive the same or similar ranks, even though just one should be considered important.

Another possibility is to use principle component analysis (PCA), which operates on just the $\bf X$ explanatory matrix. PCA transforms data into a new space characterized by eigenvectors of $\bf X$ and identifies features that explain the most variance in the new space. If the first principal component covers a large percentage of the variance, the ``loads'' associated with that component can indicate importance of features in the original $\bf X$ space. PCA is limited to linear relationships, however, and ``most variation'' is not always the same thing as ``most important.''

For classification data sets, the Relief algorithm \citep{relief} tries to identify features that distinguish between classes through repeated sampling of the data. For a sampled observation $\xi$, the algorithm finds the nearest observation with the same class (hit) and the nearest observation with the other class (miss). The score of each attribute, $x_j$, is then updated according to the distance from the selected $\xi$ to the hit and miss observations'  $x_j$ values. ReliefF \citep{ReliefF} extended Relief to work on multiclass problems and RReliefF \citep{RReliefF} adapted the technique to regression problems by ``...introduc[ing] a kind of probability that the predicted values of two instances are different.''

In an effort to deal with codependencies, data analysis techniques can rank features not just by {\em relevance} (correlation with the response variable) but also by low {\em redundancy}, the amount of information shared between codependent features, which is the idea behind minimal-redundancy-maximal-relevance (mRMR) by \citet{mRMR}. mRMR selects features in order according to the following score.

\[
J_{\text{mRMR}}(x_k) = I(x_k, y) - \frac{1}{|S|} \sum_{x_j \in S} I(x_k, x_j)
\]

\noindent where $I(x_k, x_j)$ is some measure of mutual information between $x_k$ and $x_j$, $S$ is the growing set of selected features, and $x_k$ is the candidate feature. mRMR only considers single-feature relationships with the response variable, and is limited to classification. See \cite{ubermRMR} for a recent application of mRMR at Uber Technologies.  For more on model-free feature importances, see the survey by \cite{survey}.  \citet{tsanas} suggests using Spearman's rank and not mutual information. \citet{meyer-microarray} looks for pairs of features to response variable associations as an improvement, while retaining reasonable efficiency. See \citet{filter-benchmark} for benchmarks comparing data analysis methods.

The fundamental problem faced by these data analysis techniques is that they measure relevance by the strength of the association between (typically) a single feature to response $y$, but $y$ contains the impact of all $x_j$ variables. Some analysis techniques, such as mRMR, only rank features and do not provide a numerical feature impact. Computing an appropriate association metric between categorical and numerical values also presents a challenge.

\subsection{Model-based techniques}

Turning to model-based techniques, feature importance methods are typically variations on one of two themes:  (1) tweaking a model and measuring the tweak's effect on model prediction accuracy or expected model output or (2) examining the parameters of a fitted model. The simplest approach following the first theme is {\em drop-column importance}, which defines $x_j$ importance as the difference in some accuracy metric between a model with all features (the baseline) and a model with $x_j$ removed. The model must be retrained $p$ times and highly-correlated features yield low or zero importances because codependent features cover for the dropped column.

To avoid retraining the model, $x_j$ can be permuted instead of dropped for {\em permutation importance} (\citealt{RF}). This approach is faster but can introduce nonsensical observations by permuting invalid values into records, as discussed in \cite{stopperm}; e.g., shifting a true {\tt\small pregnant} value into a male's record. Codependent features tend to share importance, at least when permutation importance is applied to RF models. To avoid nonsensical records for the RF case, \cite{rfimp} proposed a {\em conditional permutation importance} using the feature space partition created by node splitting during tree construction.  

Rather than removing or permuting entire columns of data, LIME \citep{lime} focuses on model behavior at the observation level. For an observation of interest, $\bf x$, LIME trains an interpretable linear model, on a small neighborhood of data around $\bf x$ to explain the relationship between variables and the response locally. SHAP was shown to subsume the LIME technique in \cite{shap}. 

SHAP has its roots in {\em Shapley regression values} \citep{shapley-regression} where (linear) models were trained on all possible subsets of features. Let $\hat{f}_S$ be the model trained on feature subset $x_S$ for $S \subset F = \{1, 2, .., p\}$. Each possible model pair differing in a single feature $x_j$ contributes the difference in model pair output towards the Shapley value for $x_j$. The complete Shapley value is the average model-pair difference weighted by the number of possible pairs differing in just $x_j$:
\vspace{-1mm}

\begin{equation}\label{eq:shap}
\phi_j(\hat{f},x_F) = \sum_{S \subseteq F \slash \{j\}}\
\frac{|S|!(|F|-|S|-1)!}{|F|!}\
 ( \hat{f}_{S \cup \{j\}}(x_{S \cup \{j\}}) - \hat{f}_S(x_S) )
\end{equation}\vspace{-1mm}

\noindent The SHAP importance for feature $x_j$ is the average magnitude of all $\phi_j$ values.  

To avoid training a combinatorial explosion of models with the various feature subsets, SHAP approximates $\hat{f}_S(x_S)$, with $\Ex[\hat{f}(x_{S},{\bf X}_{\slash S}') | {\bf X}_S' = x_S]$ where ${\bf X'}$ is called the {\em background set} (in ``interventional'' mode) and users can pass in, for example, a single vector with ${\bf X}_{\slash S}$ column averages or even the entire training set, ${\bf X}_{\slash S}$.  SHAP's implementation further approximates $\Ex[\hat{f}(x_{S},{\bf X}_{\slash S}') | {\bf X}_S' = x_S]$ with $\Ex[\hat{f}(x_{S},{\bf X}_{\slash S}')]$, which assumes feature independence and allows extrapolation of $\hat{f}$ to nonsensical records like permutation importance. By removing the expectation condition, the inner difference of equation \eqref{eq:shap} reduces to a function of FPDs, which means SHAP can have biased results in the presence of codependent features, as shown in \cite{stratpd}.  As implemented, then, the average SHAP value at any $x_j$ value is a point on the $FPD_j - \bar{y}$ curve and so $\overline{|FPD_j-\bar{y}|}$ = $\overline{|\phi_j(\hat{f},x)|}$. 

\cut{
\begin{figure}[htbp]
\centering
\includegraphics[scale=0.53]{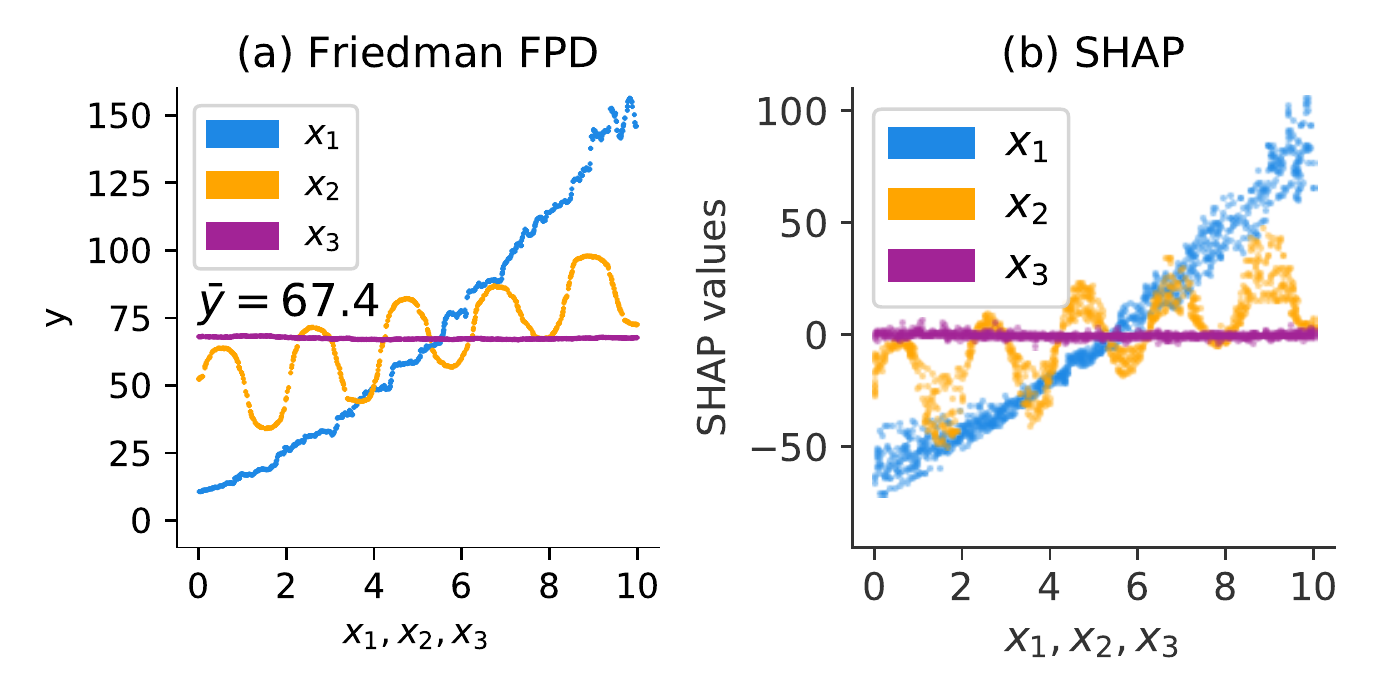}\vspace{-3mm}
\caption{\small Partial dependence plots of $n=1000$ data generated from noiseless $y = x_1^2 + x_1 x_2 + 5 x_1 sin(3 x_2) + 10$ where $x_1,x_2,x_3 \sim U(0,10)$ and $x_3$ does not affect $y$. The model is a RF with 30 trees trained on all data (training $R^2=0.997$, Out-of-bag $R^2=0.968$). SHAP used all $\bf X$ as background data.}
\label{fig:FPD_vs_SHAP}
\end{figure}

If we assume for the moment that all features are independent, there is a simple relationship between SHAP and mean-centered FPDs, the partial dependence curves as originally defined by Friedman.  \figref{fig:FPD_vs_SHAP}a and \figref{fig:FPD_vs_SHAP}b illustrate a clear similarity between FPD-$\bar{y}$ and a plot of SHAP values $(x_j^{(i)}, \phi(\hat{f},x_j^{(i)}))$. 
}

\begin{figure}[htbp]
\begin{center}
\includegraphics[scale=0.5]{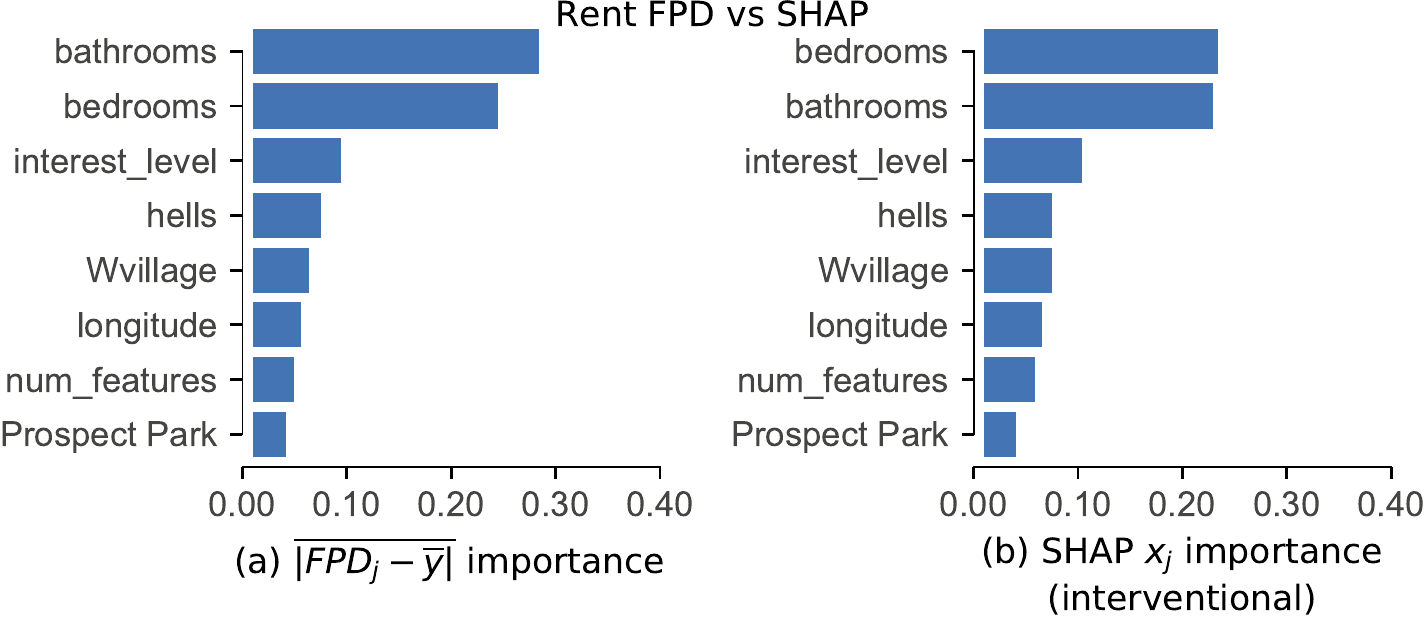}\includegraphics[scale=0.5]{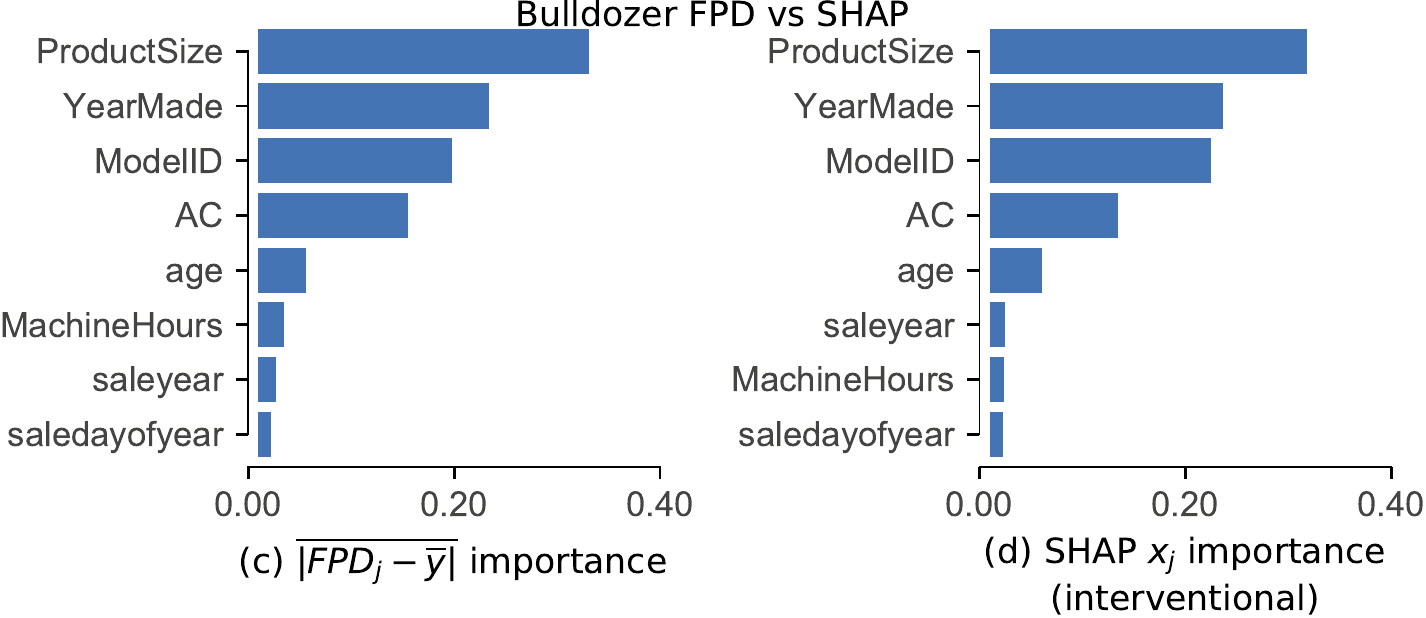}\\
\vspace{-3mm}
\caption[short]{\small  Importance ranking of top 8 features for rent and bulldozer data sets demonstrating strong similarity between average magnitude of Friedman's partial dependence curves, (a) and (c), and average magnitude of SHAP values, (b) and (d). 20,000/5,000 training/validation records were sampled to tune an RF model by 5-fold cross validation grid search. Rent validation $R^2 = 0.857$, bulldozer $R^2 = 0.856$. The FPD curve and SHAP values were computed using 300 records from the validation set; SHAP used 100 backing records from the training data.}
\label{fig:fpd_imp}
\end{center}
\end{figure}

That observation begs the question of whether measuring the area under a simple mean-centered FPD curve would be just as effective as the current SHAP implementation.  \figref{fig:fpd_imp} compares feature importances derived from FPD curves and SHAP values for two real data sets, rent from \cite{rent} and bulldozer from \cite{bulldozer}. The rank and magnitude of the feature importances are very similar between the techniques for the top  $p=8$ features. At least for these examples, the complex machinery of SHAP is unnecessary because nearly the same answer is available using a simple FPD. 

There are two methods used heavily in practice that define importance in terms of model parameters. The first operates on linear models and divides $\beta$ coefficients by their standard errors and the second examines the decision nodes in tree-based methods. The most well-known tree-based method is ``mean drop in impurity'' (also called ``gini drop'') by \cite{CART}, but there are a number of variations, such as the technique for gradient boosting machines (GBMs) described by \cite{PDP}.  The importance of $x_j$ is the average drop in $y$ impurity, entropy (classification) or variance (regression), for all nodes testing $x_j$, weighted by the fraction of test samples that reach those nodes. Such importance measures are known to be biased towards continuous or high cardinality explanatory variables (see \citealt{permbias} and \citealt{RFunbiased}).

Deriving feature importances from partial dependences as \simp{} does was previously proposed by \cite{pdvim}, and is implemented in an R package called {\tt vip}, but they defined $x_j$'s importance as ${\it PD}_j$'s ``flatness'' rather than the area under the ideal ${\it PD}_j$ curve, as we have in equation \eqref{eq:pd}.  To measure flatness, they suggest standard deviation for numerical variables and category level range divided by four for categorical variables (as an estimate of standard deviation).  Because standard deviation measures the average squared-difference from the average response, {\tt vip} would likely be more sensitive to partial dependence curve spikes than the area under the curve.  Using the idealized partial dependence, {\tt vip} would measure importance as $\overline{(PD_j - \overline{PD_j})^2}$ akin to SHAP's mass-straddling-the-mean approach, whereas we suggest the average magnitude weighted by $x_j$'s distribution.  Another difference between {\tt vip} and \simp{} is that {\tt vip} computes partial dependence curves by measuring changes in model $\hat{f}$, rather than directly from the data as we do. Any technique that computes partial dependence curves, directly or indirectly, fits neatly into {\tt vip} or the ``area under the PD curve'' framework described in this paper. FPD, ALE, SHAP, and \spd{} are four such techniques.

\citet{intgrad} introduced a technique called {\em integrated gradients} (IG) for deep learning classifiers that can also be seen as a kind of partial dependence. To attribute a classifier prediction to the elements of an input vector, $\bf x$, IG integrates over the gradient of the model output function at points along the straight-line path from a baseline vector, $\bf x'$, to $\bf x$. IG estimates the integral by averaging the gradient computed at $m$ points and multiplying by the difference between ${\bf x}'_j$ and ${\bf x}_j$:

\begin{equation}\label{eq:IG}
\text{IntegratedGradient}_j(\hat{f}, {\bf x},{\bf x'}) = (x_j - x'_j) \times \frac{1}{m} \sum_{k=1}^{m} \frac{\partial \hat{f}({\bf x}' + \frac{k}{m}({\bf x}-{\bf x}'))}{\partial x_j}
\end{equation}

\noindent  Because IG integrates the partial derivative like ALE and \spd,  equation \eqref{eq:IG} can be interpreted as the $x_j$ partial dependence curve evaluated at $\bf x$ weighted by the range in $x_j$ space. Alternatively, the average gradient within an $x_j$ range times the range of $x_j$ is equivalent to the area under the $x_j$ gradient curve in that range (by the mean value theorem for integrals). In that sense, \cite{intgrad} is also similar to \simp, except that we integrate the gradient twice, once to get the partial dependence curve and a second time to get the area under the partial dependence curve.

\simp{} is neither a model-based technique nor a model-free technique. It does not use model predictions but does rely on \spd, which internally uses a decision tree model to stratify feature space. \simp{}, therefore, has a lot in common with the ``mean drop in impurity'' technique. The  differences are that users do not provide a fitted tree-based model to \simp{} and \simp{} examines leaf observations rather than decision nodes. Unlike techniques relying on model predictions, \simp{} can compute feature impact rather than feature importance. Unlike model-free techniques (such as mRMR), \simp{} is able to provide impacts not just feature rankings and can consider the relationship between multiple features and the response.   \simp{} (via \spd) also performs well in the presence of codependent variables, unlike many model-based techniques. In the next section, we demonstrate that \simp{} is effective and efficient enough for practical use.

\section{Experimental results}\label{sec:experiments}

Assessing the quality of feature impact and importance is challenging because, even with domain expertise, humans are unreliable estimators (which is why we need data analysis algorithms in the first place).  The simplest approach is to examine impacts and importances computed from synthetic data for which the answers are known.  For real data sets, we can train a predictive model on the most impactful or most important $k$ features, as identified by the methods of interests, and then compare model prediction errors; we will refer to this as top-$k$. (\citealt{mRMR} and \citealt{tsanas} also used this approach.) The method that accurately identifies the most impactful features without getting confused by codependent features should yield lower prediction errors for a given $k$.  

In this section, we present the results of several experiments using the toy Boston data set and three real data sets: NYC rent prices \citep{rent}, bulldozer auction sales \citep{bulldozer}, and flight delays \citep{flights}. Rent has $p=20$ features, bulldozer has 14, and flight has 17. We draw $n$=25,000 samples from each data set population, except for Boston which only has 506 records, and split into 80\% training / 20\% validation sets. Each point on an error curve represents the validation set mean absolute error (MAE) for a given model, feature ranking, and data set. All models used to rank features or measure top-$k$ errors were tuned with 5-fold cross-validation grid search across a variety of hyperparameters using just the training records. (Tuned hyperparameters can be found at the start of file {\tt\small genfigs/support.py} in the repo.) The same \spd{} and \cspd{} hyperparameters were used across all simulations and datasets.\footnote{
The entire \simp{} code base is available at {\tt\small https://github.com/parrt/stratx} and running {\tt\small articles/imp/genfigs/RUNME.py} will regenerate all figures in this paper, after downloading the three Kaggle data sets.  \spd{} and \cspd{} calls always used {\tt\small min\_samples\_leaf=20}. Simulations were run on a 4.0 Ghz 32G RAM machine running OS X 10.13.6 with SHAP 0.35, scikit-learn 0.21.3, XGBoost 0.90, and Python 3.7.4. The same random seed of 1 was set for each simulation for graph reproducibility.}

\begin{figure}
\centering
\begin{subfigure}{.245\textwidth}
    \centering
\includegraphics[scale=0.43]{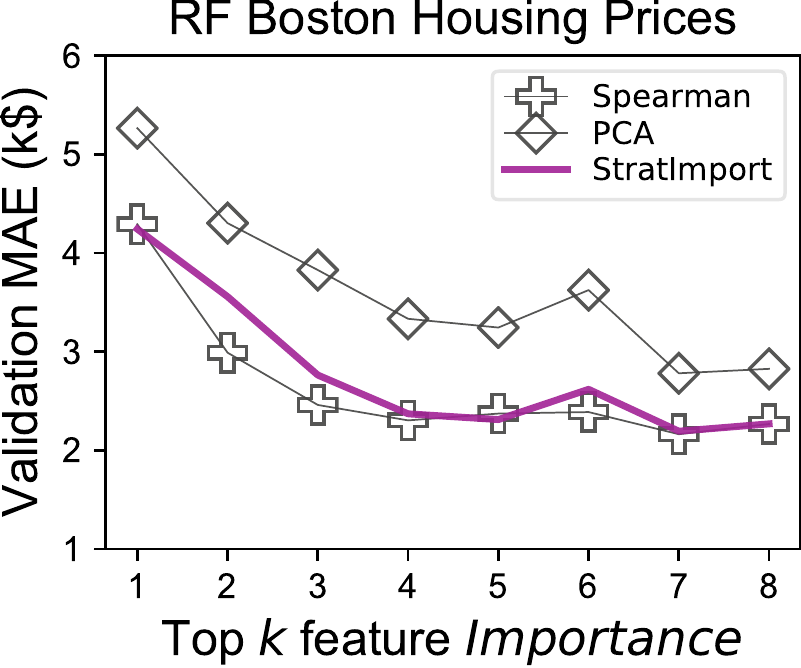}
\vspace{-2mm}
\subcaption{}
\end{subfigure}%
\begin{subfigure}{.245\textwidth}
    \centering
\includegraphics[scale=0.43]{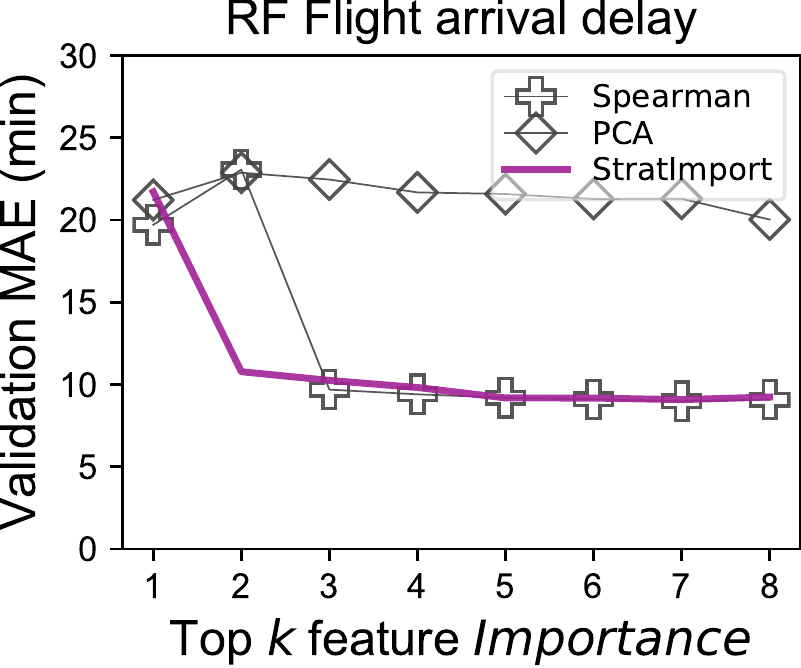}
\vspace{-2mm}
\subcaption{}
\end{subfigure}
\begin{subfigure}{.245\textwidth}
    \centering
\includegraphics[scale=0.43]{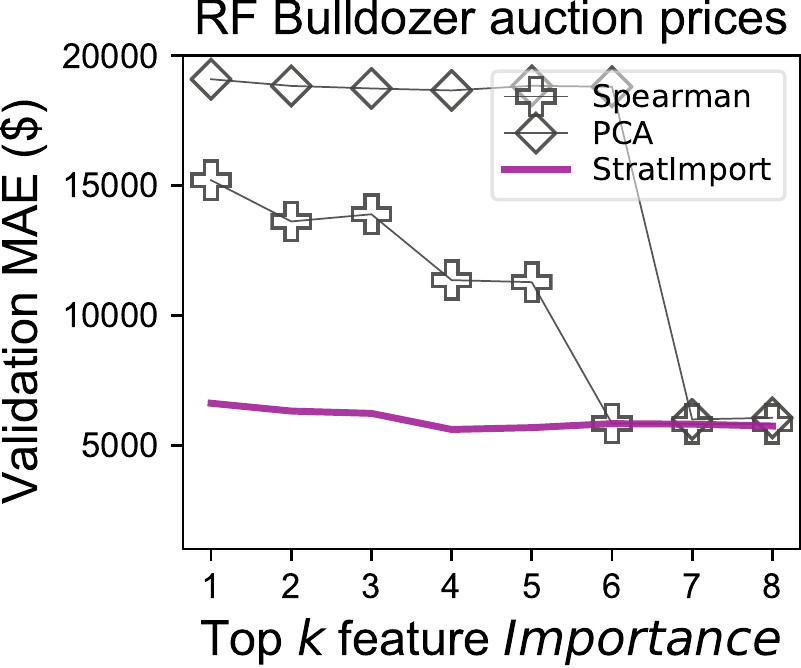}
\vspace{-2mm}
\subcaption{}
\end{subfigure}
\begin{subfigure}{.245\textwidth}
    \centering
\includegraphics[scale=0.43]{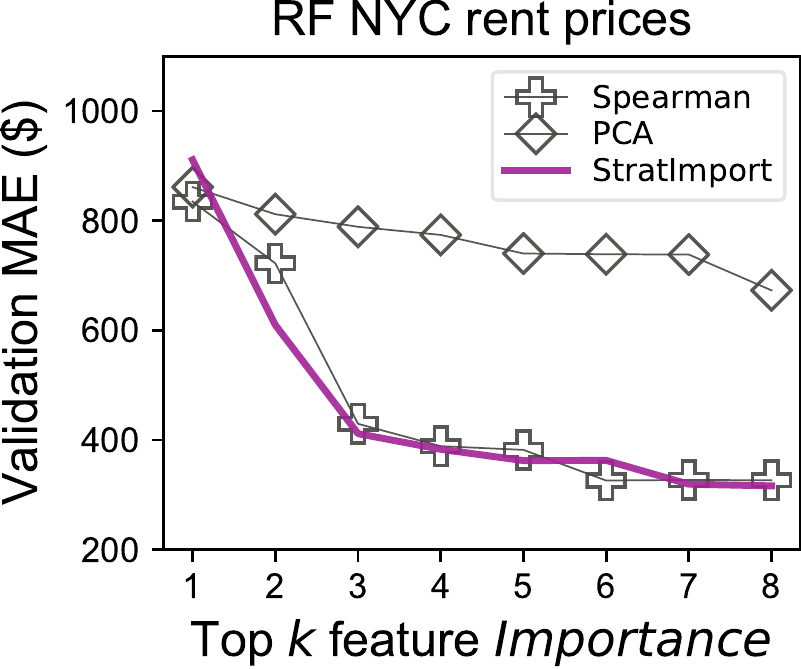}
\vspace{-2mm}
\subcaption{}
\end{subfigure} 
\vspace{-3mm}
\caption{\small {\bf RF MAE curves from \underline{baseline} rankings} computed using RF models trained on Boston, flight, bulldozer, and rent data sets. Error curves represent 5-fold cross validation using the top-$k$ features as ranked by Spearman's R, PCA ``loads'' associated with the first component, and \Impo{}.}
\label{fig:baseline}
\end{figure}

We begin with a baseline comparison of \Impo{} to principal component analysis' (PCA) ranking (``loads'' associated with the first component) and to Spearman's R coefficients computed between each $x_j$ and the response variable $y$.  \figref{fig:baseline} shows the MAE curve for an RF model trained using the top-$k$ features as ranked by PCA, Spearman, and \Impo. Spearman's R does a good job for all but the bulldozer data set, while PCA performs poorly on all four data sets. \Impo{} is competitive with or surpasses these baseline techniques. 

\cut{
\figref{fig:baseline} also shows the error curve for the features ranked by ordinary least squares (OLS); a feature's score is its $\beta$ coefficient divided by its standard error. (OLS is not applicable to the bulldozer data set because it has many high-cardinality categorical explanatory variables, which would create tens of thousands of dummy variables.) 

OLS curves are similar to \Impo's except for rent in \figref{fig:baseline}d and are included as a common reference curve on subsequent graphs.
}

\begin{figure}
\centering
\begin{subfigure}{.245\textwidth}
    \centering
\includegraphics[scale=0.45]{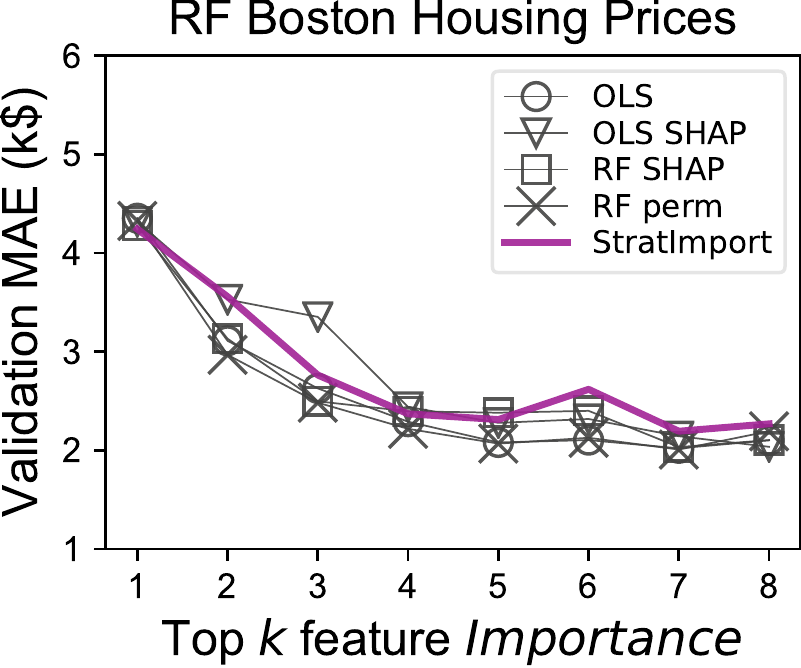}
\end{subfigure}%
\hfill
\begin{subfigure}{.245\textwidth}
    \centering
\includegraphics[scale=0.45]{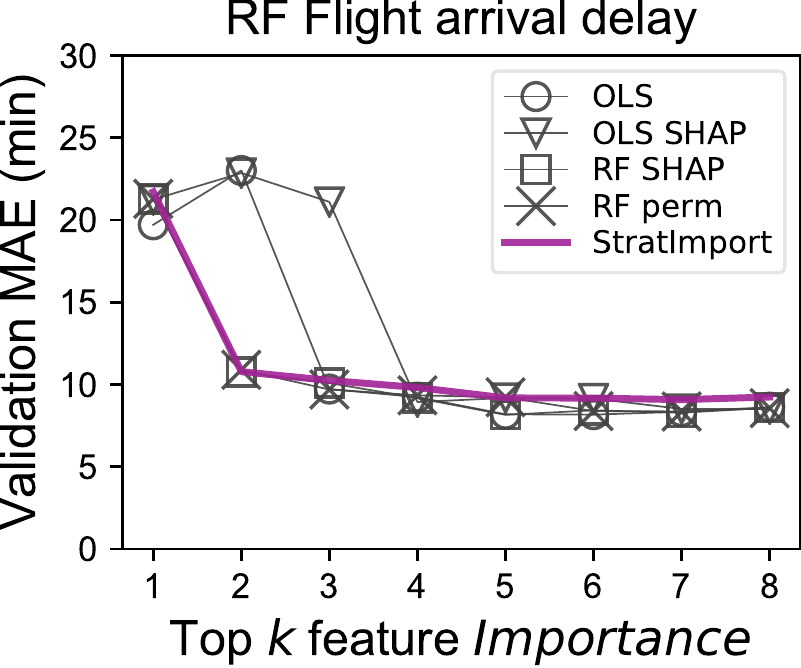}
\end{subfigure}
\hfill
\begin{subfigure}{.245\textwidth}
    \centering
\includegraphics[scale=0.45]{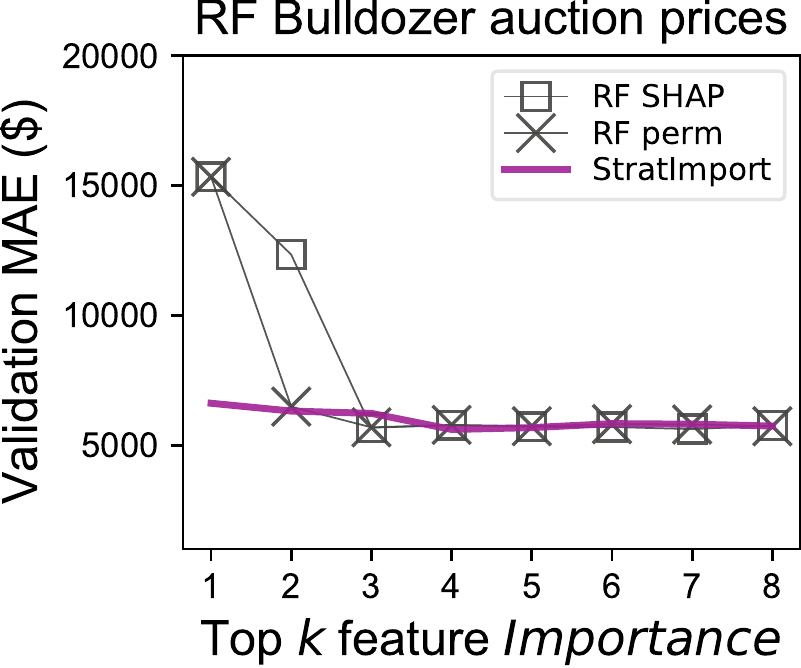}
\end{subfigure}%
\hfill
\begin{subfigure}{.245\textwidth}
    \centering
\includegraphics[scale=0.45]{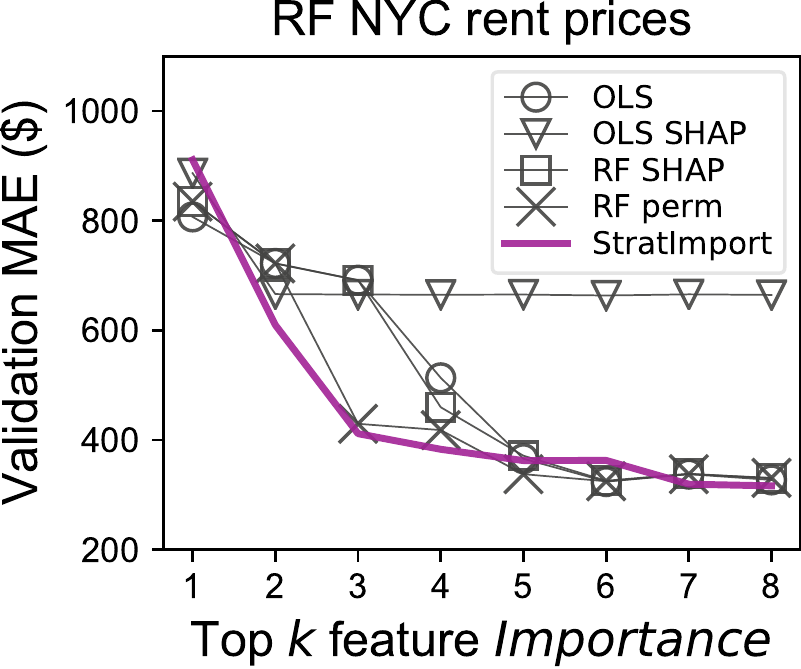}
\end{subfigure}

\vspace{1mm}
\begin{subfigure}{.245\textwidth}
    \centering
\includegraphics[scale=0.45]{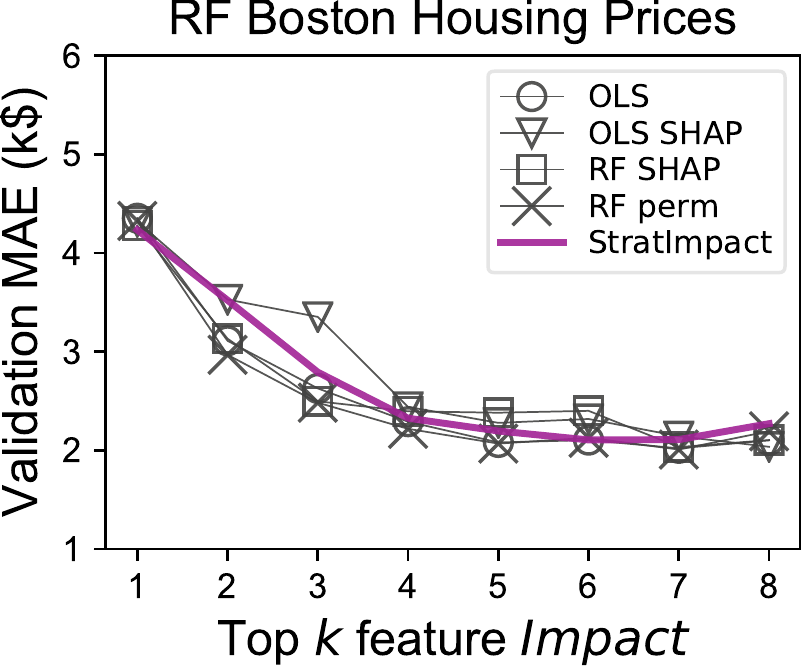}
\vspace{-5mm}
\subcaption{}
\end{subfigure}%
\hfill
\begin{subfigure}{.25\textwidth}
    \centering
\includegraphics[scale=0.45]{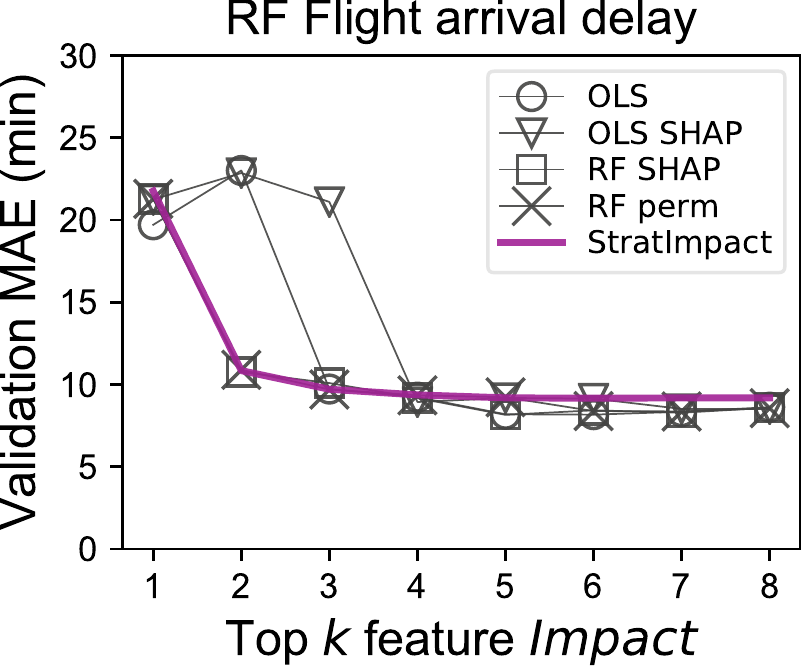}
\vspace{-5mm}
\subcaption{}
\end{subfigure}
\hfill
\begin{subfigure}{.25\textwidth}
    \centering
\includegraphics[scale=0.45]{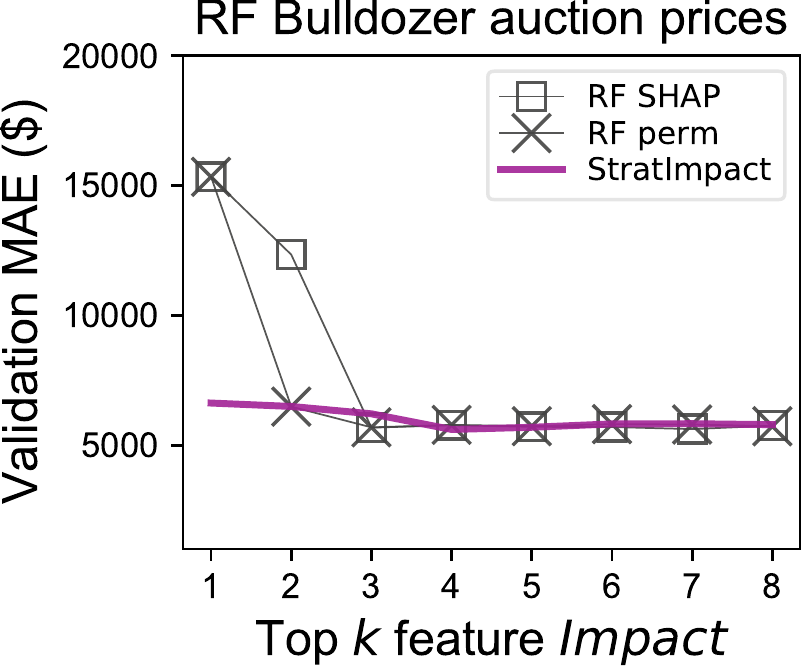}
\vspace{-5mm}
\subcaption{}
\end{subfigure}%
\hfill
\begin{subfigure}{.245\textwidth}
    \centering
\includegraphics[scale=0.45]{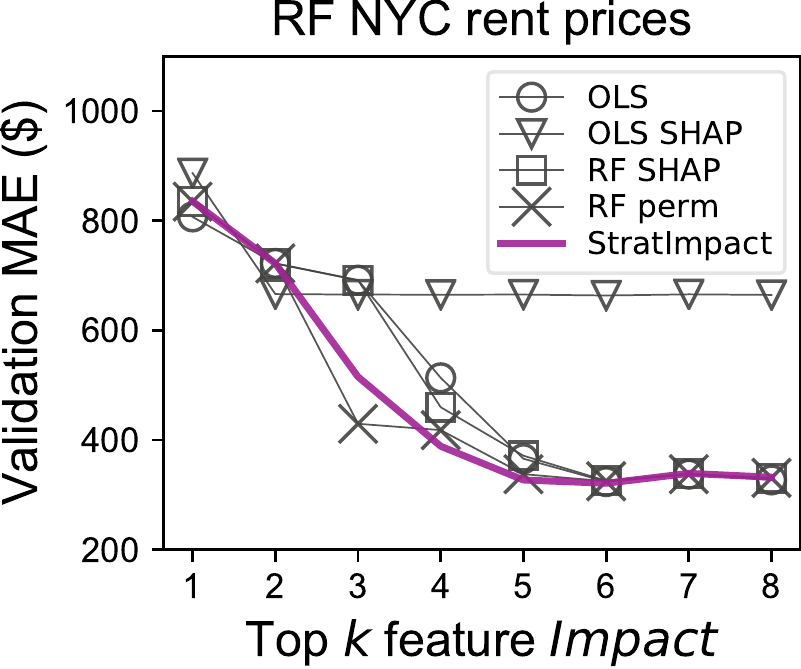}
\vspace{-5mm}
\subcaption{}
\end{subfigure}
\vspace{-3mm}
\caption[short]{\small {\bf RF MAE curves from {\bf importance} rankings on top and {\bf impact} rankings on bottom}. The mean absolute error curves from 40-tree RF models trained on boston, flight, bulldozer, and rent data sets. Error curves represent 5-fold cross validation using the top-$k$ features from the following feature rankings: OLS as in \figref{fig:baseline}, SHAP interrogating OLS, SHAP interrogating 40-tree RF, permutation importance interrogating 40-tree RF, and \simp{}. All methods had access to 80\% training data from $n=25,000$ random sample (Boston has just 506 records).  All methods except \simp{} had access to the 20\% validation set.}
\label{fig:topk}
\end{figure}

Next, in \figref{fig:topk}, we compare \Impo{}'s rankings to those of ordinary least squares (OLS), RF-based permutation importance, and SHAP interrogating OLS and RF models. A feature's OLS ranking score is its $\beta$ coefficient divided by its standard error. (OLS is not very useful for the bulldozer data set because there are important label-encoded categorical explanatory variables.) OLS and OLS SHAP analyzed all $n$ records for each data set, but RF-based permutation importance and RF-based SHAP trained on 80\% then used the remaining 20\% validation set to rank features. We deliberately restricted \simp{} to analyzing just the 80\% training data, to see how it would fare. RF-based SHAP and permutation importance, therefore, have two clear advantages: (1) they use the same kind of model (RF) for both feature ranking and for computing top-$k$ error curves and (2) they are able to select features using 100\% of the data, but \simp{} selects features using just the  training data.

The error curves for \simp, RF SHAP, and permutation importance are roughly the same for Boston in \figref{fig:topk}a and flight in \figref{fig:topk}b. For bulldozer in \figref{fig:topk}c, \simp's error curve suggests it selected a more predictive feature than RF SHAP or RF permutation importance for $k=1$: {\tt ModelID}  followed by high-cardinality categorical {\tt YearMade}. RF permutation importance chose {\tt ProductSize} as most important followed by {\tt ModelID} and RF SHAP chose {\tt ProductSize} followed by {\tt YearMade}. For the rent data set in \figref{fig:topk}d, \simp{} selects {\tt brooklynheights} (L1 distance from the apartment to that neighborhood) as the most predictive feature followed by {\tt bedrooms}.  RF SHAP and permutation importance selected {\tt bedrooms} followed by {\tt bathrooms}.  The \simp{} curve is very similar to the RF permutation curve.   The second row in \figref{fig:topk} shows that the plain impact, unweighted by $x_j$'s  distribution, can also work well for model feature selection purposes.  The impact error curves match the importance error curves closely, indicating that $x_j$ density is not critical for these data sets.

The OLS-derived feature rankings present an interesting story here (the circular markers). For Boston, flight, and rent, OLS feature rankings give error curves that are fairly similar to those of RF SHAP and RF permutation importance.  Please keep in mind that, while the features were chosen by interrogating a linear model, the error curves were  computed using predictions from a (stronger) RF model.  Perhaps most surprising is that OLS chooses the best single most-predictive feature for both flight and rent data sets, choosing {\tt TAXI\_OUT} and {\tt bathrooms}, respectively; RF SHAP and RF permutation rank {\tt DEPARTURE\_TIME} and {\tt bedrooms} as most predictive. The circle markers representing OLS in \figref{fig:topk}b and \figref{fig:topk}d have the lowest $k=1$ error curve value,  suggesting that, for example, the number of bathrooms is more predictive of rent price than bedrooms in New York. The error curve derived from the OLS SHAP feature rankings differs from the OLS curve because the OLS coefficients are divided by the standard error. An error curve using raw OLS coefficients is the same as OLS SHAP's curve.

\cut{
\begin{figure}
\centering
\begin{subfigure}{.245\textwidth}
    \centering
\includegraphics[scale=0.45]{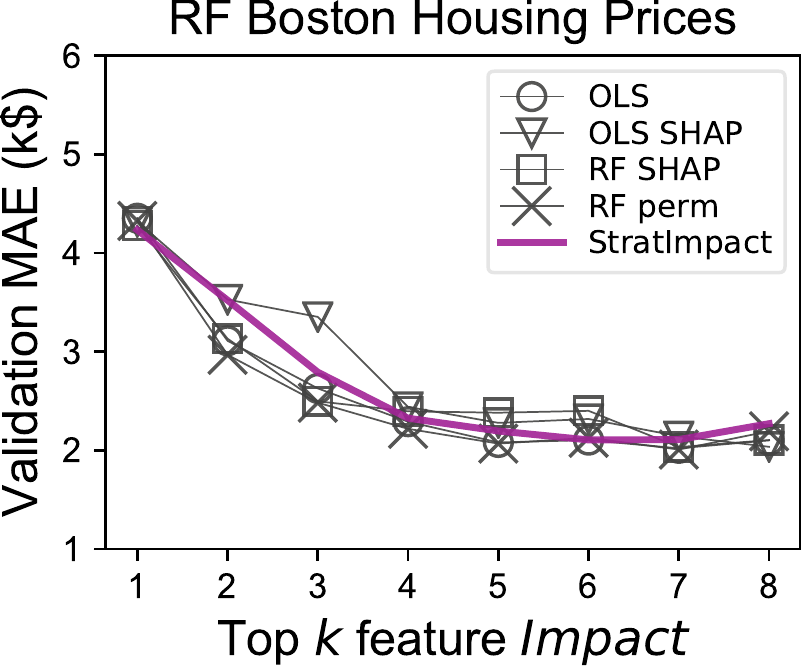}
\subcaption{}
\end{subfigure}%
\hfill
\begin{subfigure}{.245\textwidth}
    \centering
\includegraphics[scale=0.45]{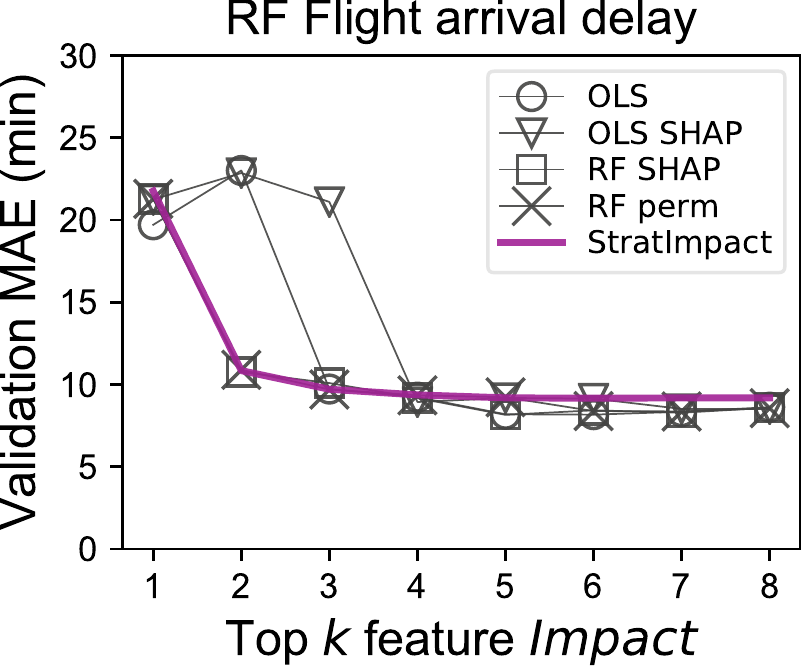}
\subcaption{}
\end{subfigure}
\hfill
\begin{subfigure}{.245\textwidth}
    \centering
\includegraphics[scale=0.45]{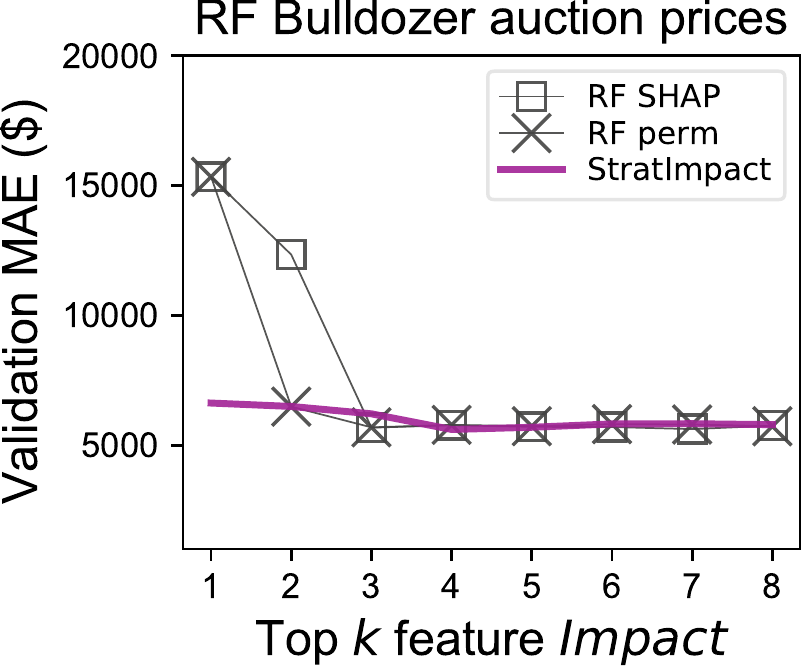}
\subcaption{}
\end{subfigure}%
\hfill
\begin{subfigure}{.245\textwidth}
    \centering
\includegraphics[scale=0.45]{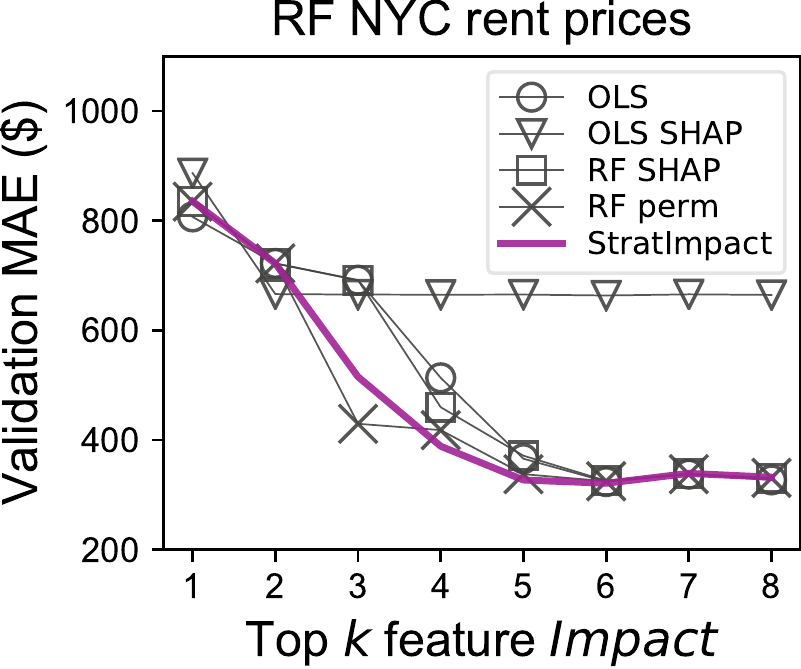}
\subcaption{}
\end{subfigure}
\caption[short]{\small MAE curves as in \figref{fig:topk} except using \simp{} {\em impact} rather than {\em importance}. \todo{merge with previous importance graphs?}}
\label{fig:topk-impact}
\end{figure}
}

\begin{figure}
\centering
\begin{subfigure}{.245\textwidth}
    \centering
\includegraphics[scale=0.45]{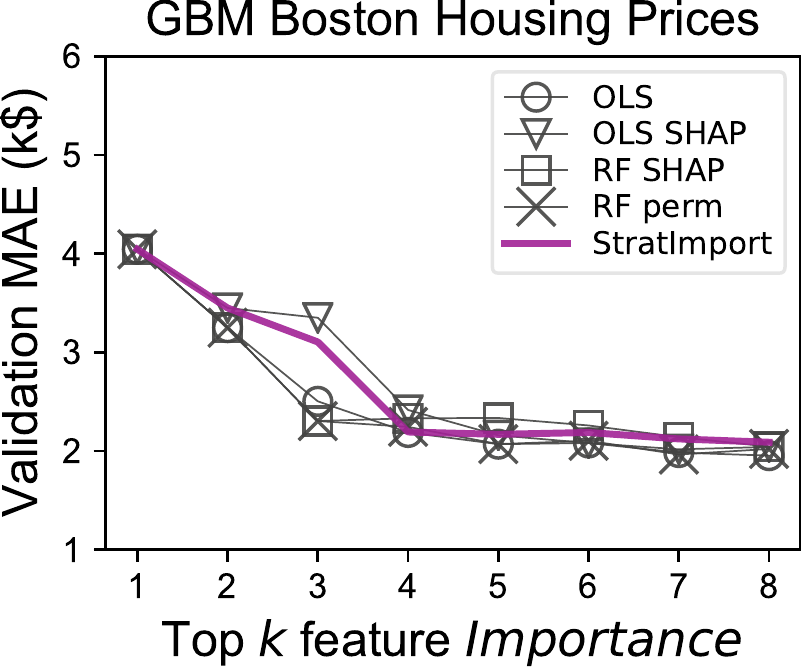}
\vspace{-5mm}
\subcaption{}
\end{subfigure}%
\hfill
\begin{subfigure}{.245\textwidth}
    \centering
\includegraphics[scale=0.45]{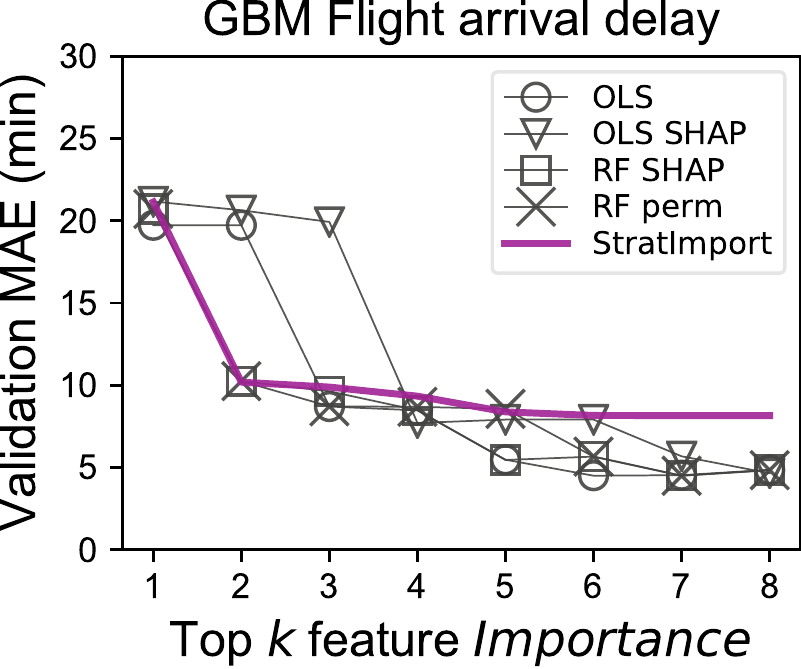}
\vspace{-5mm}
\subcaption{}
\end{subfigure}
\hfill
\begin{subfigure}{.245\textwidth}
    \centering
\includegraphics[scale=0.45]{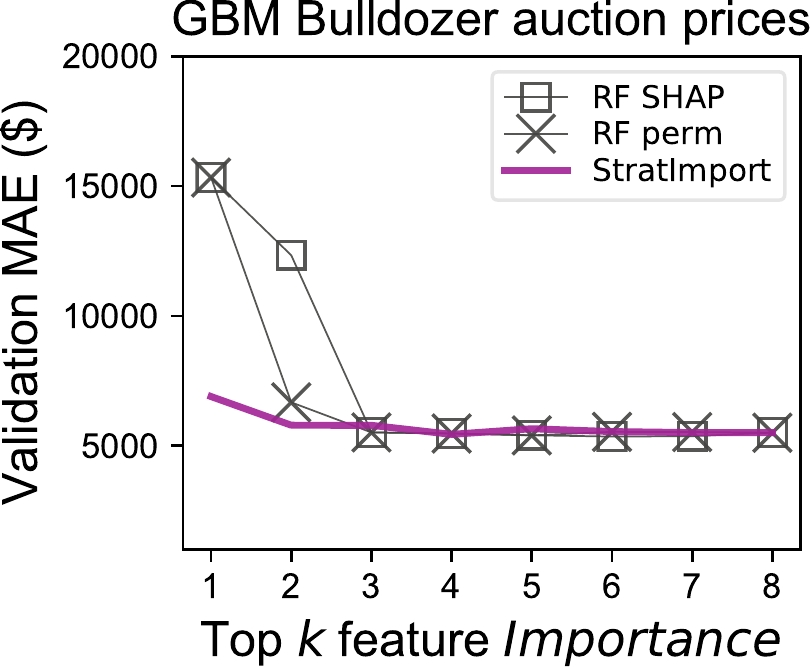}
\vspace{-5mm}
\subcaption{}
\end{subfigure}%
\hfill
\begin{subfigure}{.245\textwidth}
    \centering
\includegraphics[scale=0.45]{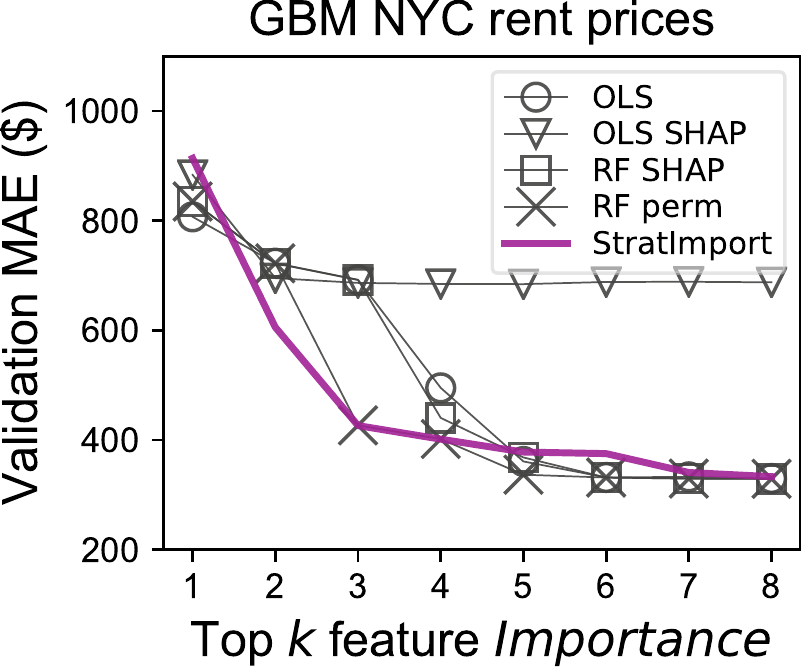}
\vspace{-5mm}
\subcaption{}
\end{subfigure}
\vspace{-3mm}
\caption[short]{\small MAE curves as in \figref{fig:topk} except measuring {\tt xgboost} predictions, rather than RF; hyperparameters were tuned via 5-fold cross validation.}
\label{fig:topk-gbm}
\end{figure}

Features that are predictive in one model are not necessarily predictive in another model.  To determine how well the feature rankings from the various methods ``export'' to another model, we trained gradient boosting machines (GBM) on the OLS-based, RF-based, and \simp{} feature rankings.  \figref{fig:topk-gbm} shows that the \simp{} error curves generated using GBM models are similar to those resulting from RF models, except for the \figref{fig:topk-gbm}(b) flight delay curve, which does not improve much after feature $k=5$, while the other feature rankings do improve. 

\begin{figure}[htbp]
\begin{center}
\includegraphics[scale=0.5]{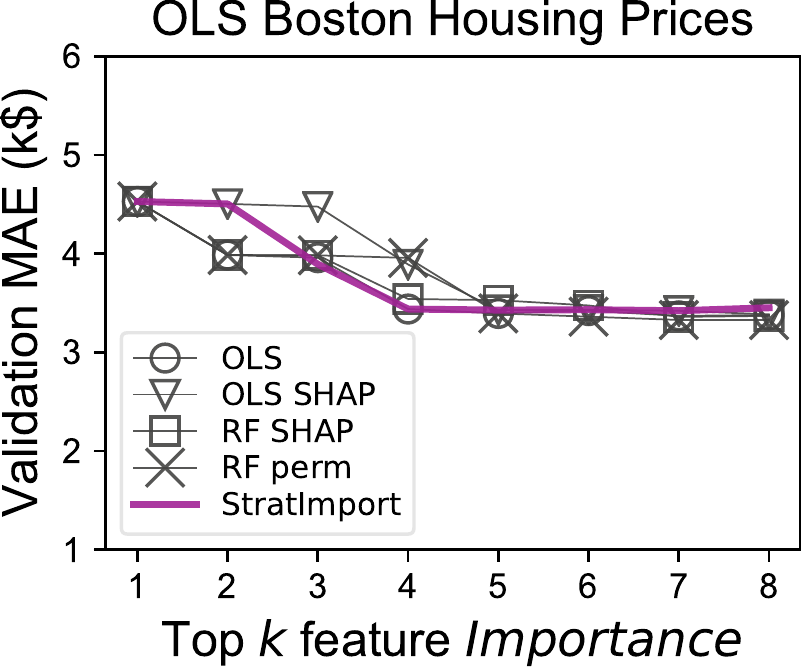}~~~
\includegraphics[scale=0.5]{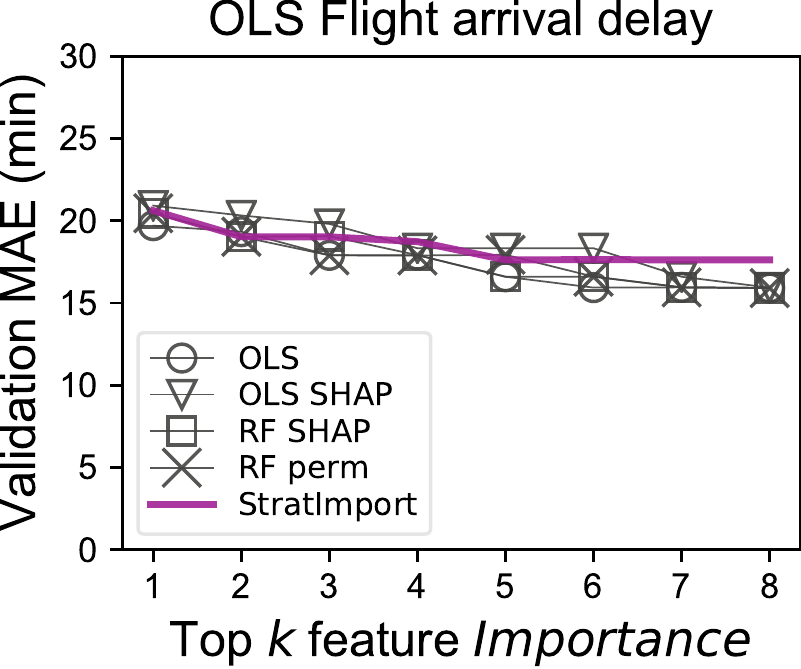}~~~\includegraphics[scale=0.5]{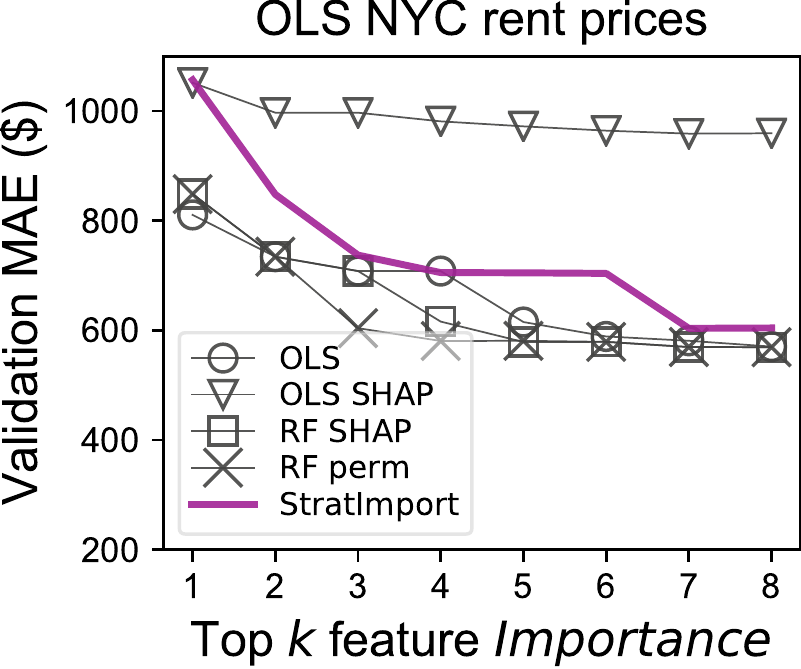}
\vspace{-3mm}
\caption{\small MAE curves as in \figref{fig:topk-gbm} but measuring OLS rather than RF predictions.}
\label{fig:OLS}
\end{center}
\end{figure}

To check feature exportation to a very different model, we computed error curves for the Boston, flight, and rent data sets using OLS regressors, as shown in \figref{fig:OLS}, again using the \simp, OLS-derived, and RF-derived rankings.  The error curves derived from OLS model predictions are all higher than those from RFs, as expected since OLS is weaker than GBM, but feature rankings from all technique export reasonably well, with the exception of \simp's rankings for rent.  The poor performance could be due to OLS' inability to leverage the \simp{} ranking or could be poor feature ranking on \simp's part.  The former is more likely, given that more sophisticated methods get low error curves using \simp's ranking.

To compute the \simp{} feature rankings described thus far, we used a single 80/20 training and validation sample in order to get reproducible figures.  But, different subsets of the data set can yield very different impacts, depending on the variability of the data set. \simp{} supports multiple trials via bootstrapping or subsampling on the $(\bf X, y)$ data set to obtain impact and importance standard deviations.  As an example, \figref{fig:stability} shows the  feature rankings of \Imp{} and \Impo{}, averaged from 30 trials using 75\% subsamples selected randomly from the complete (cleaned) 49,352 records of the rent data set. The error bars show two standard deviations above and below the average (a red error bar indicates it reaches 0 and was sifted to the bottom).  The high variances of the neighborhood features, such as {\tt brooklynheights} and {\tt Evillage}, arise because many distance-to-neighborhood features have similar impacts; selection will depend on vagaries of the subsample.
 
\setcounter{figure}{7} 
\begin{figure}
\begin{subfigure}{.49\textwidth}
    \centering
\includegraphics[scale=0.6]{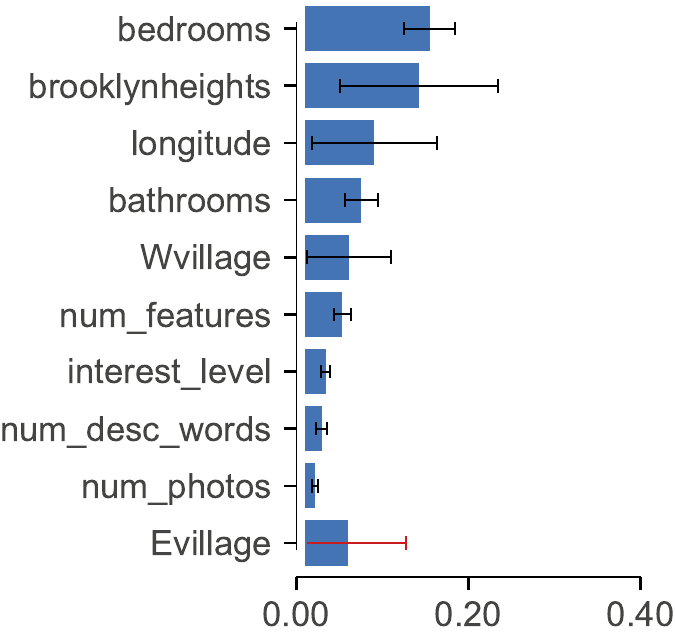}
\vspace{-2mm}
\subcaption{\simp{} Rent Importance}
\end{subfigure}%
\hfill
\begin{subfigure}{.49\textwidth}
    \centering
\includegraphics[scale=0.6]{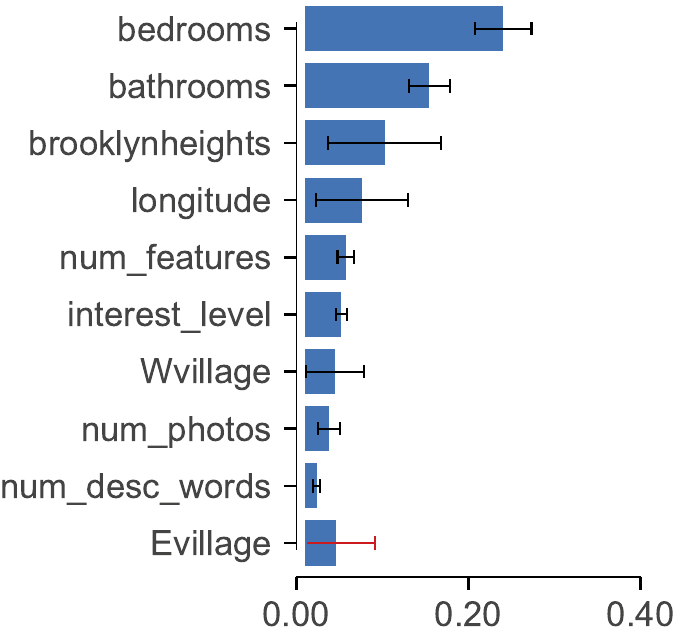}
\vspace{-2mm}
\subcaption{\simp{} Rent Impact}
\end{subfigure}%
\vspace{-3mm}
\captionof{figure}{\small Average feature importance and impact across 30 trials on the rent data set, with 75\% subsamples from 49,352 total records.  Error bars show two standard deviations. Red indicates two standard deviations reach zero.}
\label{fig:stability}
\end{figure}

Performance is important for practitioners so it is worthwhile analyzing \simp{} time complexity and demonstrating that it operates in a reasonable amount of time.  The cost of computing each $x_j$'s impact is dominated by the cost of computing the partial dependence but includes a pass over the $x_j$ values to compute the mean. Computing the importance costs an extra pass through the data to get a histogram. The upper bound  complexity for both \spd{} and \cspd{} partial dependences is $O(n^2)$ in the worst case, which is  similar to FPD's $O(nm)$ for $n$ records and $m$ curve evaluation points.  In practice, \spd{} typically performs linearly while \cspd{} exhibits mildly quadratic behavior.  The overall worst case behavior for \simp{} is then $O(p n^2)$ to get $p$ impacts and importances.   For comparison purposes, ALE is the most efficient at $O(n)$ per feature and SHAP has the hardest time with efficiency due to the combinatorial problem of feature subsetting. SHAP has model-type-dependent optimizations for linear regression, deep learning, and decision-tree based models, but other models are prohibitively expensive. For example, SHAP applied to a support vector machine trained on the 80\% Boston training set takes four minutes to explain the 101 records in the 20\% validation set.

\begin{table}\small
\centering
\begin{tabular}{r r r r r r r r r}
{\bf dataset} & $p$ & catvars & {\small $n$=1,000} & {\small 10,000} & {\small 20,000} & {\small 30,000} & time versus $n$~~ & $R^2$\\
\hline
{\tt\small flight} & 17 & 6 & 5.7s & 8.9s & 35.5s & 76.0s & {\small $-0.360 n + 0.095 n^2$} & {\small 0.9945}\\
{\tt\small bulldozer} & 14 & 3 & 0.8s & 3.0s & 11.4s & 24.6s & {\small $-0.063 n + 0.029 n^2$} & {\small 0.9961}\\
{\tt\small rent} & 20 & 0 & 0.4s & 4.0s & 8.5s & 12.9s & {\small $0.424 n + 0.000 n^2$} & {\small 0.9995}\\
\end{tabular}
\vspace{-3mm}
\caption{\small  Execution time for subsets of size 1,000 to 30,000 for rent, bulldozer, flight data sets.  There are a total of 40 numerical and 9 categorical variables: rent, bulldozer, and flight have $p=20$, $p=14$, and $p=17$. Rent has no categorical variables and exhibits linear performance, whereas categorical variables introduce mildly quadratic behavior. Final two columns describe how data fits to quadratic equations. Time does not include Numba just-in-time compiler warm-up, but users do experience this warm-up time.}
\label{fig:timing}
\end{table}

\tblref{fig:timing} summarizes the empirical time in seconds to compute importances for a range of subset sizes for the three Kaggle data sets. The ``time versus $n$'' and $R^2$ columns describe a quadratic fit to the curve representing the time (in seconds) required to compute subsets from $n=1$ to 30,000 stepping by 1,000. The rent data set has no categorical variables and grows linearly with $n$. The flight data set, on the other hand, shows quadratic behavior due to the (six) categorical variables. Despite the worst-case complexity, \tblref{fig:timing} suggests our Python-only \simp{} prototype is fast enough for use on real data sets of sizes in the tens of thousands. When the cost of training and tuning a model is counted, \simp{} would likely outperform other techniques.

Each of the $p$ feature impact computations is independent and could proceed in parallel. Unfortunately, our casual attempts at parallelizing the algorithm across multiple CPU cores was thwarted by Python's dreaded ``global interpreter lock'' (threading) or passing data-passing costs between processes (process-based threading).  We did, however, get increased performance using the Numba just-in-time compiler ({\tt\small http://numba.pydata.org}) on algorithm hotspots (at the cost of 5 seconds of compiler warmup time at runtime).

\section{Discussion and future work}\label{sec:discussion}

In this paper, we propose a nonparametric approach to measuring numerical and categorical feature $x_j$'s impact upon the response variable, $y$, based upon a mathematical definition of impact: the area under $x_j$'s ideal $PD_j$ curve (numerical) or mean-centered $PD_j$ curve (categorical). By weighting $x_j$'s  partial dependence curve with $x_j$'s distribution, we arrive at a mathematical definition of feature importance.  Our goal is not to claim that existing feature selection techniques are incorrect or fail in practice; they very often work well. Instead, we hope to:

\begin{enumerate}
\item Bring attention to the fact that feature importance is not the same as feature impact, because of potential distortions from peering through model $\hat{f}$. The same importance algorithm applied to the same data can get meaningfully different answers from different fitted models.   For the purpose of gaining insights into the behavior of objects under consideration (such as customers or patients) or the impact of their features, the ideal impact metric would avoid $\hat{f}$ predictions and operate directly on the data.\vspace{2mm}

\item Demonstrate it is possible to compute impacts without relying on predictions from a fitted model, by estimating partial derivatives of the unknown generator function $f$. This approach is valuable because there are large prospective industrial and scientific communities that lack the expertise to choose and tune machine learning models. Model-free importance techniques do exist, but they usually provide just a ranking, rather than meaningful impact values, and often consider associations of the response with just one or two features at a time.
\end{enumerate}
 
To assess the quality of \simp's feature impacts, we compared \simp's recommended feature importance rankings to those of other techniques as a proxy. Despite not having access to predictions from a fitted model nor access to the entire data set, \simp{} feature rankings are competitive with other rankings from other commonly-used techniques on Boston and three real data sets.  \figref{fig:topk}c illustrates a case in which \simp's most important feature choice yields half the error rate of the top features recommended by the other methods.  One would expect model-based techniques to easily identify the single most important model feature, or at least to do so more readily than an approach not using model predictions.  Even if such results are rare, misidentifying this top feature indicates that there is room for improvement in the feature importance research area. 

An interesting and unanticipated result is that simple expedient approaches, such as ranking features by Spearman's R coefficient or permutation importance, perform well at least for the first eight important features in our experiments. (We did not perform experiments on the least important features.) Also, the error curves for SHAP and permutation importance generally mirror each other for the first eight features.  Both techniques introduce potentially nonsensical records to avoid retraining models, but this does not appear to affect their ability to rank features for feature selection purposes.

Our proposed approach relies on accurate partial dependences, and considerable effort has gone into refining the \spd{} and \cspd{} partial dependence algorithms currently used by \simp.  Any improvement in the accuracy of estimation techniques for partial dependence curves would be useful in their own right and particularly helpful for \simp.  The current prototype is limited to regression and so, next, we hope to develop suitable model-free partial dependence and impact algorithms for classification.

\cut{
\begin{figure}
\centering
\begin{subfigure}{1\textwidth}
    \centering
\includegraphics[scale=0.5]{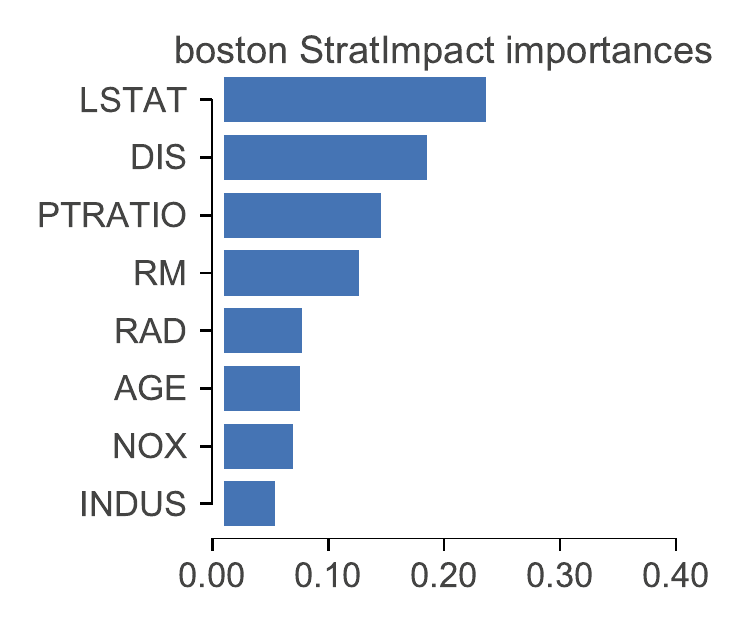}
\includegraphics[scale=0.5]{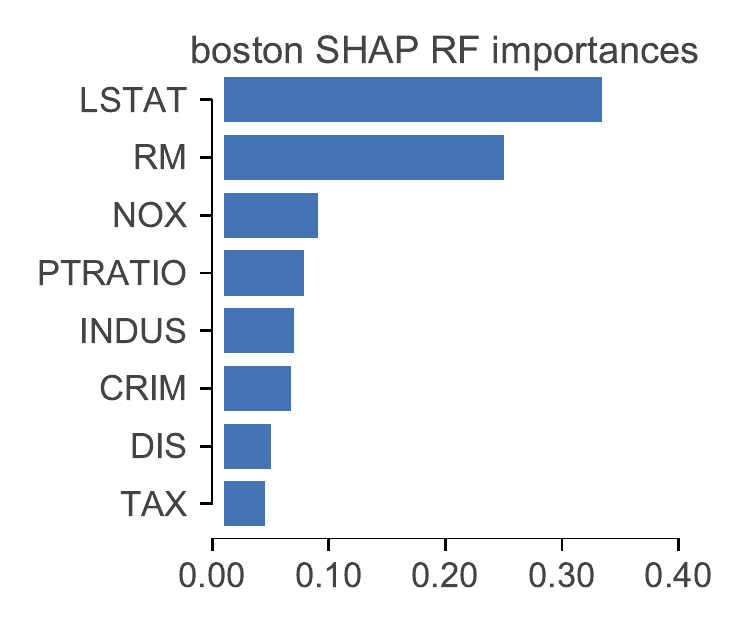}
\vspace{-2mm}\subcaption{\footnotesize Note LSTAT/RM order is diff than in original figure as their is high variance}\vspace{3mm}
\end{subfigure}%
\hfill
\begin{subfigure}{1\textwidth}
    \centering
\includegraphics[scale=0.5]{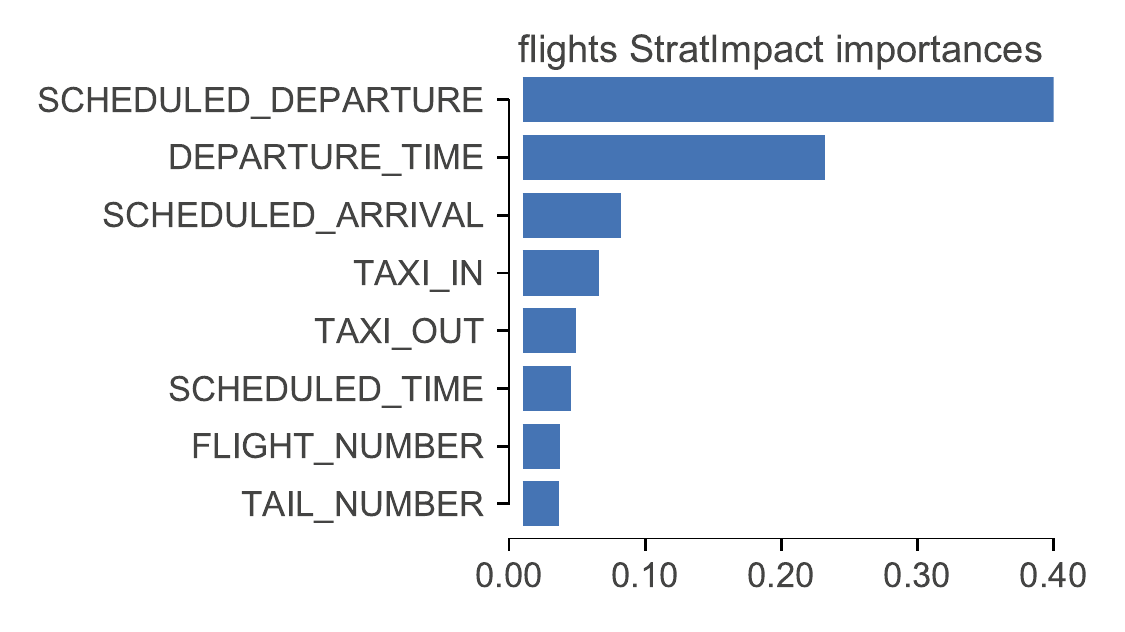}
\includegraphics[scale=0.5]{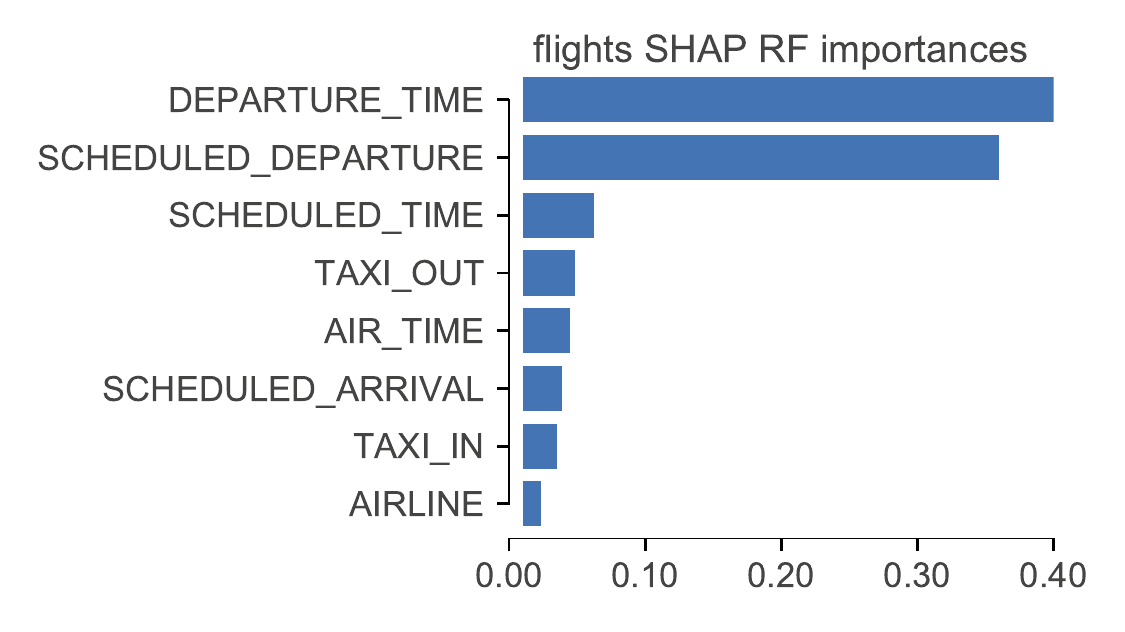}
\vspace{-2mm}\subcaption{\footnotesize 5.8M records}\vspace{3mm}
\end{subfigure}
\hfill
\begin{subfigure}{1\textwidth}
    \centering
\includegraphics[scale=0.5]{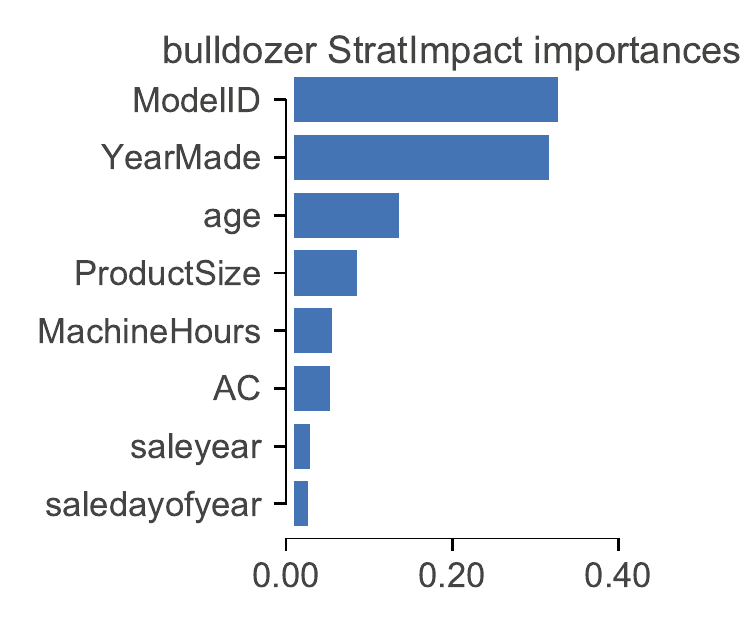}
\includegraphics[scale=0.5]{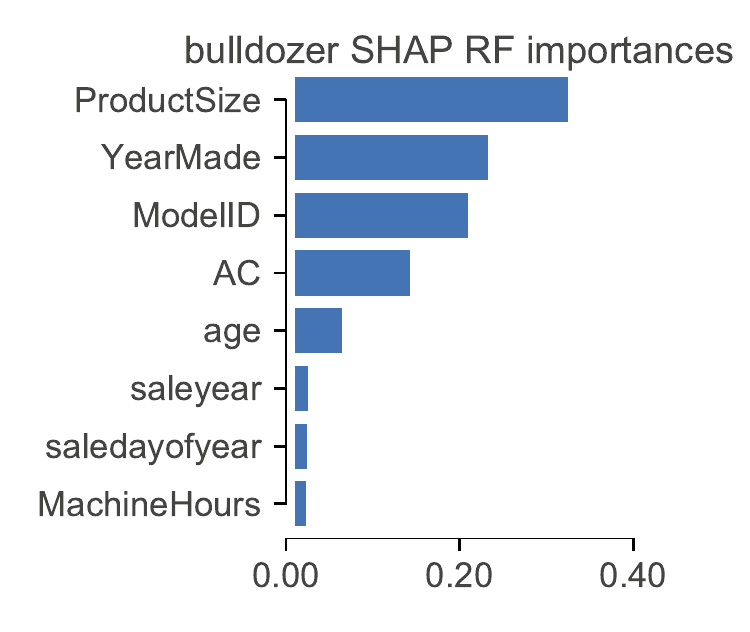}
\vspace{-2mm}\subcaption{\footnotesize foo}\vspace{3mm}
\end{subfigure}%
\hfill
\begin{subfigure}{1\textwidth}
    \centering
\includegraphics[scale=0.5]{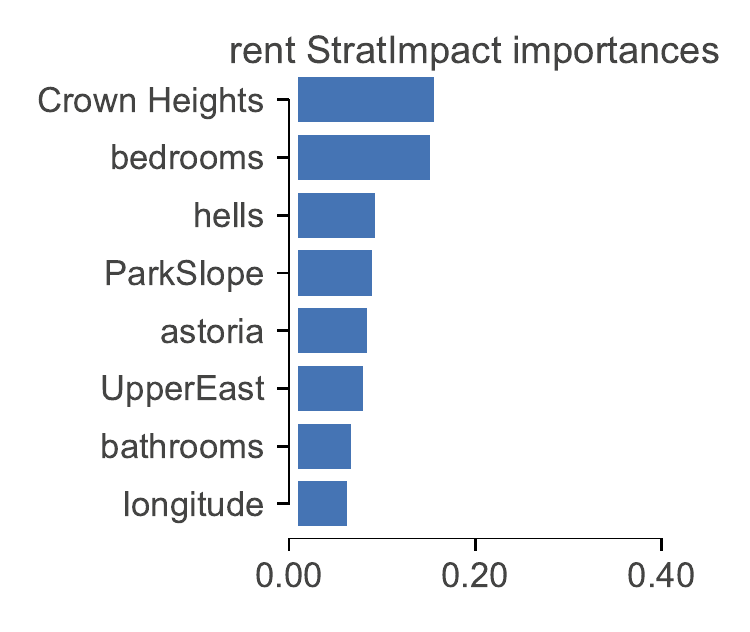}
\includegraphics[scale=0.5]{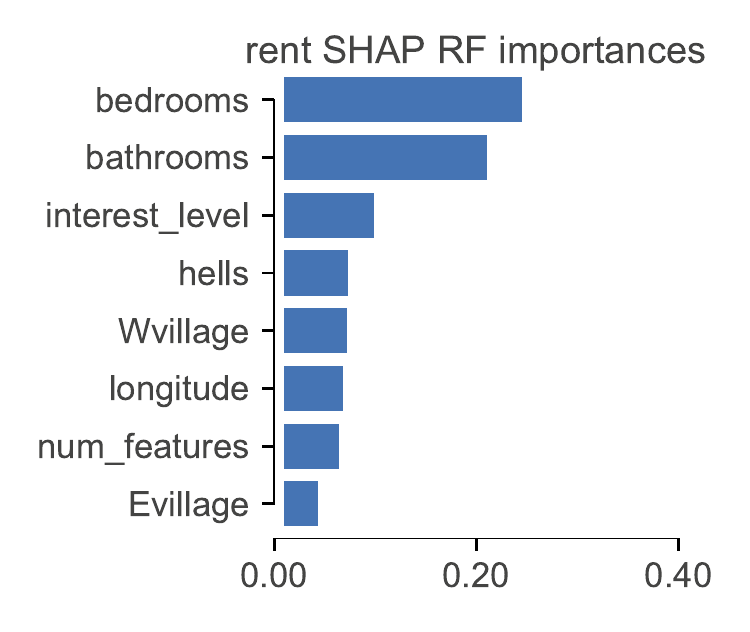}
\vspace{-2mm}\subcaption{\footnotesize foo}\vspace{3mm}
\end{subfigure}
\caption[short]{blorttttt}
\label{fig:features}
\end{figure}
}

\bibliography{pdimp}
\end{document}